\documentclass{article} 
\usepackage{collas2025_conference,times}
\usepackage{easyReview}

\usepackage{hyperref}
\hypersetup{
    colorlinks=true,
    linkcolor=red,
    filecolor=magenta,
    urlcolor=blue,
    citecolor=purple,
    pdftitle={Overleaf Example},
    pdfpagemode=FullScreen,
    }

\usepackage[normalem]{ulem}
\usepackage{booktabs}       
\usepackage{amsfonts}       
\usepackage{nicefrac}       
\usepackage{microtype}      
\usepackage{lipsum}

\usepackage{ragged2e} 
\usepackage{subcaption}
\usepackage{booktabs} 

\usepackage{amsmath}
\usepackage{amssymb}
\usepackage{mathtools}
\usepackage{amsthm}

\newif\ifcomments

\usepackage[capitalize,noabbrev]{cleveref}

\creflabelformat{equation}{#2\textup{#1}#3}

\usepackage{float}
\usepackage{graphicx}

\newif\ifcomments
\commentstrue

\theoremstyle{plain}
\newtheorem{theorem}{Theorem}[section]
\newtheorem{proposition}[theorem]{Proposition}
\newtheorem{lemma}[theorem]{Lemma}
\newtheorem{claim}[theorem]{Claim}

\theoremstyle{definition}
\newtheorem{definition}[theorem]{Definition}
\newtheorem{assumption}[theorem]{Assumption}
\theoremstyle{remark}
\newtheorem{remark}[theorem]{Remark}

\newenvironment{proofof}[1]{\noindent\textbf{Proof of {#1}}
  \hspace*{1em}}{\qed\bigskip\\}

\let\originalleft\left
\let\originalright\right
\renewcommand{\left}{\mathopen{}\mathclose\bgroup\originalleft}
\renewcommand{\right}{\aftergroup\egroup\originalright}

\def\reals{\mathbb{R}} 

\def\<{\left\langle} 
\def\>{\right\rangle}

\def\defeq{\vcentcolon=} 

\newcommand{\paren}[1]{\left(#1\right)}
\newcommand{\braces}[1]{\left\{#1\right\}}
\newcommand{\brackets}[1]{\left [ #1 \right ]}

\newcommand{\id}{\mathbf{I}} 
\def\norm#1{\left\|{#1}\right\|} 
\newcommand{\twonorm}[1]{\norm{#1}_2} 
\newcommand{\opnorm}[1]{\norm{#1}_{\text{op}}} 

\def\sp#1{\mathrm{span}(#1)}
\newcommand{\rank}[1]{\mathrm{rank}(#1)}

\def\bfv{\vv}
\def\bfu{\vu}

\def\bfa{\va}
\def\bfp{\vp}

\newcommand{\eig}[2]{\lambda_{#1}\paren{{#2}}}

\DeclareMathOperator*{\E}{\mathbb{E}} 

\def\bigO#1{\mathcal{O}\left(#1\right)} 
\DeclareMathOperator{\Tr}{Tr} 

\newcommand{\normal}{\sf{N}}

\newcommand{\lspan}{\mathrm{span}}

\def\Xmem{\mX^{\text{mem}}}
\def\Ymem{\vy^{\text{mem}}}

\def\ranj{J}

\def\Ptil{\Tilde{\mP}}
\def\Pbar{\Bar{\mP}}
\def\Xtil{\Tilde{X}}

\def\Ztil{\Tilde{\rmZ}}
\def\Pitil{\Tilde{\mPi}}

\def\Zpr{{\rmZ'}}

\def\wstar{{\vw^*}}

\usepackage{amsmath,amsfonts,bm}

\def\eqref#1{equation~\ref{#1}}

\def\1{\bm{1}}

\def\rvy{{\mathbf{y}}}

\def\rmX{{\mathbf{X}}}

\def\rmZ{{\mathbf{Z}}}

\def\va{{\bm{a}}}

\def\vp{{\bm{p}}}

\def\vu{{\bm{u}}}
\def\vv{{\bm{v}}}
\def\vw{{\bm{w}}}
\def\vx{{\bm{x}}}
\def\vy{{\bm{y}}}

\def\mA{{\bm{A}}}

\def\mP{{\bm{P}}}

\def\mW{{\bm{W}}}
\def\mX{{\bm{X}}}

\def\mPi{{\bm{\Pi}}}

\DeclareMathAlphabet{\mathsfit}{\encodingdefault}{\sfdefault}{m}{sl}
\SetMathAlphabet{\mathsfit}{bold}{\encodingdefault}{\sfdefault}{bx}{n}

\def\sW{{\mathbb{W}}}

\title{Replay can provably increase forgetting}

\author{
Yasaman Mahdaviyeh \\
  Columbia University \\
  \texttt{yasamanmdv@cs.columbia.edu}
  \And
James Lucas \\
  NVIDIA \\
  \texttt{jalucas@nvidia.com} 
   \And
   Mengye Ren \\
  New York University\\
  \texttt{mengye@nyu.edu}  
  \And
  Andreas S. Tolias \\
  Stanford University \\
  \texttt{tolias@stanford.edu} 
  \And
  Richard Zemel\\
  Columbia University \\
  \texttt{zemel@cs.columbia.edu} 
  \And
  Toniann Pitassi \\
  Columbia University \\
  \texttt{toni@cs.columbia.edu} \\
}

\preprintcopy 

\begin{document}

\maketitle

\renewcommand{\thefootnote}{\relax}

\footnotetext{
To appear in the Proceedings of the Conference on Lifelong Learning Agents (CoLLAs) 2025.
}

\begin{abstract}
Continual learning seeks to enable machine learning systems to solve an increasing corpus of tasks sequentially. A critical challenge for continual learning is forgetting, where the performance on previously learned tasks decreases as new tasks are introduced. One of the commonly used techniques to mitigate forgetting, sample replay, has been shown empirically to reduce forgetting by retaining some examples from old tasks and including them in new training episodes.
In this work, we provide a theoretical analysis of sample replay in an over-parameterized continual linear regression setting, where each task is given by a linear subspace and with enough replay samples, one would be able to eliminate forgetting. Our analysis focuses on sample replay
and highlights the role of the replayed samples and the relationship between task subspaces.
Surprisingly, we find that, even in a noiseless setting, forgetting can be non-monotonic with respect to the number of replay samples.
We present tasks where replay can be {\it harmful} with respect to worst-case settings, and also in distributional settings where replay of randomly selected samples increases forgetting in expectation.
We also give empirical evidence that harmful replay is not limited to training with linear models by showing similar behavior for a neural networks equipped with SGD. 
Through experiments on a commonly used benchmark, we provide additional evidence that, even in seemingly benign scenarios, performance of the replay heavily depends on the choice of replay samples and the relationship between tasks.

\end{abstract}
\section{Introduction}
\label{sec:introduction}
Humans and other animals can seemingly learn new skills and accumulate knowledge throughout their lifetimes. 
Continual learning algorithms aim to achieve this same capability: to produce systems that can learn from a sequence of tasks.
One of the main challenges is  a phenomenon typically  
called catastrophic forgetting~\citep{mccloskey1989catastrophic}, where the learner's performance on a previously visited task degrades once it learns new tasks.
A dominant theme in continual learning has been the development of methods
to address catastrophic forgetting \citep{li2017learning, zenke2017continual,rebuffi2017icarl,kirkpatrick2017overcoming,GDUMB}. However, there has been limited theoretical treatment of their efficacy.

In this work, we focus on a continual learning problem consisting of a sequence of linear regression tasks. The tasks are designed such that a single linear model is sufficient to solve the full sequence. 
The aim of focusing on this problem is that
it is simple enough to permit analysis while preserving some key challenges of the continual learning problem.
A particularly noteworthy discovery of prior work is that even in this setting, catastrophic forgetting can occur~\citep{evron22a}. 
However, an open question remains: Can methods designed to combat forgetting succeed in this setting? As a first step towards answering this question, we study one such method, experience replay.

There are many variations of experience replay. For example, \citet{BrainGR} introduce a brain-inspired form of generative replay and show that it has strong performance on complex benchmarks.
In this paper, we focus on the more standard form of sample replay, an intuitive technique where samples observed during continual learning are stored and repeated back to the learner during later tasks, to help retain solutions to prior tasks. This simple method has been shown
to be effective at ameliorating forgetting~\citep{rebuffi2017icarl,rolnick2019experience,aljundi2019gradient,wu2019large,Chaudhry2019OnTE,tiwari2022gcr}, and is often used in dynamic learning settings like reinforcement learning~\citep{lin1992reinforcement}.
Replay has also been a focus of study in neuroscience, as strong experimental evidence supports the hypothesis that replay plays an important role in memory consolidation \citep{RaschBorn,Oudiette2013}.

Some existing theoretical works in continual learning show that when tasks are repeatedly visited in a cyclical order, forgetting will be minimized. \citet{evron22a} shows this in the continual linear regression setting, while \citet{Chen2022MemoryBF} show a similar result in a more general PAC-like continual learning setting. 
While these results show that forgetting decreases when the entire task sequence is replayed, they do not consider what happens when replay occurs between tasks, and involves a subset of samples.
This is a focus of this work.

We first prove that somewhat counter-intuitively there are worst-case scenarios where replay actually causes more forgetting. We then prove a surprising stronger result, that this can still hold in a certain average case sense: 
even when examples are sampled randomly from specific task subspaces, and replay samples are chosen randomly, replay can increase forgetting on average.
We emphasize that we are obtaining these results in a relatively benign (or non-adversarial) setting: the samples are not noisy, and the tasks share an optimal solution. 
This increased forgetting occurs even when the tasks are close to each other, under a natural notion of distance.

In addition to our theoretical contributions, we provide an empirical investigation of forgetting with sample replay to support our theoretical findings. We verify our theoretical results and further show that the same surprising behavior exists in continual linear regression learning with neural networks. Through experiments on MNIST continual learning benchmarks, we show that there is significant variation in the effectiveness of replay, and 
find a simple task sequence where replay can increase forgetting.

\section{Background and Setup}
\label{sec:background and Setup}
\subsection{Background}\label{sec: background}
\cite{evron22a} initiated the study of catastrophic forgetting in overparameterized linear regression. 
They consider a sequence of linear tasks $\paren{\mX_t, \vy_t}_{t=1}^T$ where $\mX_t \in \reals^{n_t \times d}$, $\vy_t \in \reals^{n_t}$ and $n_t, d$ are the number of samples per task and input dimension respectively. 
They assume that the sequence of linear tasks share a solution that could be obtained by jointly training on all tasks, and for each task $n_t < d$, so any single task would not necessarily contain all the information needed to learn the common solution.
Despite the existence of a common solution, they show that there are sequences of tasks such that learning them in a sequential manner with gradient descent will result in a significant amount of forgetting, which is defined to be the average error on all previously seen tasks (\cref{def: forgetting}). 
In this setting, $d$ many samples would be sufficient to recover the solution to each task. 

Another group of techniques used to mitigate forgetting is through regularization. 
For example, forgetting can be eliminated using a Fisher information based weighting matrix
\citep{kirkpatrick2017overcoming,evron2023classification}. We note that storing such matrices would take order $d^2$ bits of memory, which is of the same order as storing $d$ samples.
\citet{ICL} introduce a general notion of an "ideal continual learner" and instantiate it for continual linear regression.
Their algorithm maintains the shared null space of the previous tasks.
Again, storing a null space could take order $d^2$ bits of memory.
Sample replay, on the other hand, allows the learner to store much less than $d$ many samples, reducing memory requirements significantly.
However, the effectiveness of replaying a few samples in this linear setting is an open question and the focus of this paper.

\subsection{ General Setup} \label{sec: setup}
Here we consider two  different settings for sample replay: the worst case and the average case.
We first introduce the general setup,
which is shared between the two settings and closely resembles the earlier formulation
described above, and then examine each separately, in Sections \ref{sec: worst case setup} and \ref{sec: average case setup}, respectively.
We start with the following two assumptions.
\begin{assumption}[Over-parameterized linear regression]\label{assump: overparameterized lr}
    We assume that each task is:
\begin{itemize}
    \item Linear: there is $\vw^*_t \in \reals^d$ such that $ \mX_t \vw^*_t  = \vy_t$.
    \item Over-parameterized:  $k_t \defeq \rank{\mX_t} < d$.
\end{itemize}
\end{assumption}

We also assume 
realizability, 
which ensures that the $T$ tasks share a common solution:
\begin{assumption}\label{assump: realizability}
(Realizability). There exists $\vw^* \in \bigcup_t \lspan(\mX_t)$, where $\twonorm{\vw^*} \le 1$,  such that for all tasks $t$, 
$\vy_t = \mX_t \vw^*$.
\end{assumption}
\paragraph{Projections.}
Let $\mPi_t$ be an orthogonal projection onto the row span of $\mX_t$, i.e. the span of the samples of task $t$.
We can write $\mPi_i = \mX_t^+ \mX_t$ where $\mX_t^+$ is the Moore-Penrose inverse of  $\mX_t^+$. 
Another way to obtain the orthonormal projection  is using a matrix $\mW_t$ whose columns form an orthonormal basis for the row span of $\mX_t$. 
Given $\mW_t$, we could write $\mPi_t = \mW_t \mW_t^\top$.
We use $\mP_t \defeq  \id - \mPi_i$ to denote the orthogonal projection onto the null space of task $t$. 
We use the term projection interchangeably with orthogonal projection throughout the rest of this paper.

\paragraph{Learning Procedure.}
Initially $\vw_0$ is set to the all-zero vector. For each task $t$, starting with the solution $\vw_{t-1}$ from the previous task(s), the learning algorithm minimizes squared error $\twonorm{\mX_t \vw - \vy_t}^2$ using GD or SGD.

It is known that training with GD or SGD leads to a solution that has minimum distance to initialization \citep{gunasekar18a,rethinking_generalization}, that is, 
\begin{align} \label{eq: learning procedure}
    \vw_t = \arg \min_{\vw} \twonorm{\vw - \vw_{t-1}} \;\; \mbox{s.t.} \; \mX_t \vw = \vy_t.
\end{align}
The {\it parameter error} of the procedure in \cref{eq: learning procedure} satisfies the following recursive relationship
\begin{align}\label{eq: residual error}
    \vw_t - \vw^* = \mP_t (\vw_{t-1} - \vw^*).
\end{align}
We include a derivation of this relationship in 
\cref{sec: derivation of recursive error} for completeness.
Initially, the parameter error vector is $\vw_0 - \vw^* = -\vw^*$, \cref{eq: residual error} states that after training on task $t$, the parameter error vector is projected onto the null space of task $t$. So the parameter error vector evolves as a sequence of orthonormal projections into task null spaces, while forgetting also takes into account projection of the parameter error onto the task samples.
\begin{definition}[Forgetting] \label{def: forgetting}
Given a sequence of training samples for tasks $S = \paren{(\mX_t, \vy_t)}_{t=1}^T$, the forgetting with respect to the training samples is defined to be 
\begin{equation} 
    F_S(\vw_T) = \frac{1}{T-1} \sum_{t = 1}^{T-1} \twonorm{\mX_t \vw_T - \vy_t}^2.
\end{equation}
\end{definition}

We drop the subscript $S$ when it is clear from the context which sequence of tasks the forgetting is being computed over. 
Note that the average forgetting defined above is over $T-1$ tasks since forgetting on the last task is always zero. 
 We can consider forgetting for a certain task ordering catastrophic, when $\lim_{T \rightarrow \infty} F_S(\vw_T) > 0$, or in other words, when it does not vanish with the number of tasks.

\begin{remark}
    In our average case result, each task is given by a distribution, and forgetting is measured on new samples from previous tasks' distributions. We introduce and discuss these details in \cref{sec: average case setup}.
\end{remark}
Forgetting for the output of the learning procedure described in \cref{eq: learning procedure} can be written as 
\begin{align}\label{eq: forgetting sequential projections}
    F_S(\vw_T) = & \frac{1}{T-1} \sum_{t = 1}^{T-1} \twonorm{\mX_t (\vw_T - \vw^*)}^2
     =  \frac{1}{T-1} \sum_{t = 1}^{T-1} \twonorm{\mX_t \mP_T P_{T-1} \dots P_1 \vw^*}^2,
\end{align}
see \cref{sec: derivation of basic forgetting} for this derivation.
We can see from \cref{eq: forgetting sequential projections} that forgetting not only depends on the parameter error vector but also on its 
relationship with the training samples. 
\paragraph{Replay.}
We consider a simple and standard formulation of replay in the literature,  where the learning algorithm can store up to $m$ samples from the previously seen tasks in memory, sometimes called episodic memory \citep{Chaudhry2019OnTE}. Let $\{\Xmem, \Ymem\}$ denote the set of stored samples. 
During training on the current task $t$, in addition to the current task's samples, the model also trains on $\{\Xmem, \Ymem\}$ to get the new iterate $\Tilde{w}_{t+1}$. There are many ways the algorithm can update $\{\Xmem, \Ymem\}$ \citep{Chaudhry2019OnTE}. In our worst case setup, this choice is adversarial, while in the average case setup we consider a random selection.

\section{Replay Can Provably Increase Forgetting in Continual Linear Regression}
\label{sec:theory}

In this section we show that replay can increase forgetting in two different settings.  Each setting demonstrates a different scenario where replay can increase forgetting.
The worst case setting highlights how the relationships between individual samples could lead to catastrophic forgetting with replay, while the average case result goes beyond interactions between individual samples and highlights the role of task subspaces and the angle(s) between them.
 For both of these results, interference within samples of each task plays an important role. Since samples within a task can be revisited many times during training, this intra-task sample interference does not matter without replay and we only see forgetting due to interference across tasks. With replay, however, since only a fraction of the samples within a task are trained on, intra-task sample interference could also contribute to forgetting.

\subsection{Worst Case: From Vanishing to Catastrophic Forgetting via Replay Sample Selection}\label{sec: worst case setup}
Our result in the worst case setting shows that the increase in forgetting due to replay can be significant. 
In addition to Assumptions \ref{assump: overparameterized lr} and \ref{assump: realizability}, we require that samples have unit norm:
\begin{assumption}\label{assump: sample norm worst case}
    Let $\mX_{ti}$ be the $i$th row of $\mX_t$. Assume that $\twonorm{\mX_{ti}} = 1$.
\end{assumption}
This assumption is not necessary to get the worst case result. 
We have included it to make it explicit that the construction in the worst case does not rely on some samples have a much larger norm than the rest.

\subsubsection{Worst case result}
In this worst case setting, we choose the tasks and a (possibly empty) subset of samples from each task to be replayed. In this setup, we show that forgetting could increase from vanishing, $\bigO {\frac{1}{T}}$, to $\Theta(1)$, which 
we have labeled catastrophic.

At the core of the tasks constructed in the worst case setting are three samples $\vx_1,\vx_2$ and $\vx_3$, where $\vx_1$ and $\vx_3$ are orthogonal to one another, and $\vx_2$ is linearly independent but not orthogonal to $\vx_1$ or $\vx_3$. Consider a sequence of two tasks where the samples for the first task consists of $\vx_1,\vx_2$, and the second task contains only $\vx_3$. After training on the first task, there is no error on $\vx_1$ and $\vx_2$. When we train on the second task, since $\vx_3$ is not orthogonal to $\vx_2$, training on $\vx_3$ introduces some error on $\vx_2$ but doesn't introduce error on $\vx_1$, since it is othogonal to $\vx_3$. 
Now suppose that $\vx_2$ is replayed, so for the second task, we train on both $\vx_2$ and $\vx_3$. Since $\vx_1$ is not orthogonal to $\vx_2$, this will introduce error on $\vx_1$. Thus in this three sample, two task setup, replay causes an {\it exchange} of the error on $\vx_2$ with error on $\vx_1$. See  \cref{fig: worst case intuition } in \cref{appendix: worst case intuition} for a geometric illustration of this phenomena.

\begin{theorem}[Worst case replay]\label{thm: worst case counter example}
    Under assumptions \ref{assump: overparameterized lr} \ref{assump: realizability} and \ref{assump: sample norm worst case}, for any $T \ge 2, d \ge 3$, there is a sequence of $T$ tasks and a sample $(\Tilde{\vx}, \Tilde{y})$ such that without replay, forgetting is $F(\vw_T) = \bigO {\frac{1}{T}}$, while with replay of  $(\Tilde{\vx}, \Tilde{y})$, forgetting is catastrophic, i.e.,  $F(\Tilde{\vw}_T) =  \Theta(1)$ .
\end{theorem}
In the proof, which is given in \cref{proof: worst case theorem} , we construct a scenario where this type of error exchange  is detrimental. 
We describe a sequence of tasks $t=1,\ldots, T$, where the sample $\vx_1$ occurs in all but the last task, while the sample $\vx_2$ occurs in just one of these tasks. Here without replay there is no forgetting on any sample other than $\vx_2$.
Also, replaying $\vx_1$ would not change the final iterate $w_T$, since without replay, there is no error on $\vx_1$.
However, if $\vx_2$ is replayed in the last task ($t=T$), causing error on $\vx_1$, the forgetting will be much larger, growing with the number of tasks $T$, since $\vx_1$ occurs in almost all the tasks.

\subsection{Average Case: Forgetting in a Random Sample Setting} \label{sec: average case results}
In the previous section, we saw that increased forgetting can arise from adversarially chosen replay samples.
Next we will show that there are task subspaces such that replay increases forgetting, even when task samples are chosen randomly from those subspaces and replay samples are also chosen randomly from previous tasks. 
On the other hand, in \cref{sec: replay cannot increase forgetting} we give some conditions under which replay cannot increase forgetting.

\subsubsection{Average Case Setup} \label{sec: average case setup}

The average case setup is more natural and general relative to the worst case setup. Each task's samples are drawn from a distribution supported on some subspace, and the forgetting is measured on a set of new samples from that distribution. Additionally, the replay samples are chosen randomly. 

In the average case construction, the task distributions are 
Gaussians supported on specific subspaces. 
Each of these subspaces can be specified by an orthonormal basis $\mW_t \in \reals^{d \times k_t}$.
The rows of the $n_t \times d$ dimensional matrix $\rmX_t$ consist of individual samples $\rmX_{t1}, \dots, \rmX_{t n_t}$, where each sample
\begin{align} \label{eq: avg case sample generation}
    \rmX_{tj} = \mW_t \rmZ_{tj} \;\;  \mathrm{and} \; \; \rmZ_{tj} \sim \normal (0, \frac{\id_{k_t}}{k_t}),
\end{align}
 and $\rmZ_{tj}$ are iid for $j \in [n_t]$.
Since $\rmX_{tj}$ are rows of $\rmX_t$, we can write $\rmX_t = \rmZ_t \mW_t^\top$, where 
$\rmZ_t$ is a $n_t \times k_t$ dimensional matrix whose rows are $\rmZ_{tj}$.
Then $\vw^*$ along with $\rmX_t$ determines $\vy_t =  \rmX_t \vw^*$.

In this average case setting, we need to have enough samples for each task to span each task subspace $\mW_t$. The sample generation process described above ensures that this condition is met as long as the number of samples for each task is larger than the task's rank.

\begin{assumption}\label{assump: avg case subspace covered}
    Assume that the number of the samples for each task is at least as large as the rank of the given subspace $\mW_t$, that is, $k_t \le n_t $.
    
\end{assumption}

While it will always be the case that $\rank{\rmX_t} \le n_t$, the condition above is to ensure that $\rank{\rmX_t} = k_t $.
We consider the expectation of forgetting with respect to $k_t$ test samples from previous tasks. To mark this difference we use $\rmX'_t, \rvy'_t $ to denote the test samples and $F_{S'}(\vw_T)$ to denote forgetting with respect to these new samples. That is
\begin{align}
    F_{S'}(\vw_T) = \frac{1}{T-1} \sum_{t=1}^{T-1} \twonorm{ \rmX'_t \vw_T - \rvy'_t }^2 = 
    \frac{1}{T-1} \sum_{t=1}^{T-1} \twonorm{\rmX'_t (\vw_T - \vw^*)}^2.
\end{align}
We assume that there are $k_t$ test samples. This would ensure that there are enough test samples to span the task's subspace.  
Next proposition gives the expected forgetting (without replay) in this setting, the proof is given in \cref{proof: expected forgetting without replay}.

\begin{proposition}\label{prop: expected forgetting without replay}
    Suppose that $\rmX_{tj}$ are sampled according to \cref{eq: avg case sample generation}, then 
    \begin{align} \label{eq: expected forgetting}
         \E \brackets {F_{S'}(\vw_T)}
    = & \frac{1}{T-1} \sum_{t = 1}^{T-1}  \twonorm{\mPi_t \mP_T \dots \mP_1 \vw^*}^2.
    \end{align}
\end{proposition}

\

\subsubsection{Average Case Results}
We now present one of our main results 
which states that replay can increase forgetting even in an average-case replay setting.

\begin{theorem} \label{thm: average case high dimensional construction}
Suppose that Assumptions \ref{assump: overparameterized lr}, \ref{assump: overparameterized lr}, and \ref{assump: realizability} hold.
Then for any $\wstar \in \reals^d$, there exists constants $ 0 < c_1, c_2, c_3 $ such that for $c_1 < d$, $ c_2 m < d-1$ and $d-1 < \frac{\exp(m \log m)}{c_3}$, there 
 is a sequences of two tasks such that  replay of $m$ randomly chosen samples from the first task increases forgetting in expectation. 
\end{theorem}

This result shows that there can be task sequences where most choices of sample options for replay are unfavorable, so that it is not only choices of replay samples that matters but also the relationship between the tasks. 
Note that replay could increase forgetting even when the number of replay samples is linear in the dimension.
The proof of \cref{thm: average case high dimensional construction} is given in \cref{proof: average case high dimensional}. Note the statement in \cref{thm: average case high dimensional construction} concerns the {\it expected} forgetting with replay; it does not imply that replaying any particular sample from the first subspace would increase forgetting. In fact, there are  directions in the first task's subspace such that replaying a sample in those directions would reduce forgetting.

Now we give an overview of our average case construction, and the intuition behind it. 
Fix an orthonormal bases  $\bfv_1, \dots, \bfv_d$ for $\reals^d$. 
Suppose that the first task's null space is spanned by a vector $\bfp_2$ that is very close, but not equal to $\bfv_2$. 
The second task's null space is $d-1$ dimensional and  spanned by $\braces{\bfv_1, \bfv_2, \dots, \bfv_{d-1}}$.
It is known that forgetting depends on the angle between the task null spaces in a non-monotonic way; see \cref{sec: explanation for angle between null spaces} for more details. 
Without replay, this would be the angle between the $\bfp_2$ and $\bfv_2$. 
Replay would reduce the second task's subspace to a subset of it, determined by the replay samples.
We show that the projections of the replayed samples into the second task's null space approximately follows a Gaussian distribution. This means that for sufficiently large $d$,
the span of the projected samples will not be too close to $\bfv_2$.
Therefore, the remaining subspace, which would be the null space of the second task with replay, would be close to $\bfv_2$, but not fully include $\bfv_2$. 
Consequently, the angle with $\bfp_1$ would increase, but not by too much, becoming closer to $\pi/4$, resulting in increased forgetting.

Note that the above Theorem holds when the underlying dimension $d$ is a sufficiently large. However, the result holds even when the dimension $d$ is very small, but with a somewhat more complicated argument. In  \cref{thm: avg case counter example} (see \cref{proof: thm avg case}) we
prove a similar result when $d=3$.

\subsection{When Replay Cannot Increase Forgetting} \label{sec: replay cannot increase forgetting}
These results naturally raise the question of when replay is guaranteed to be benign, i.e. not result in higher forgetting relative to no replay. 
We give some simple sufficient conditions for benign replay.
\begin{itemize}
    \item If forgetting without replay is zero. 
    In our setting where after training the loss on each task is zero, zero forgetting implies that loss on all previous tasks is zero.
    If loss on all previous tasks is zero, then loss on any set of selected replay samples is zero. This means that, the gradient of the loss with respect to those samples would also be zero, resulting in no change to the solution.
    This holds regardless of whether we are in the average case setting or the worst case one.
    It also holds for longer sequences of tasks.
    The results in \citet{evron22a} give some hints on when forgetting would be zero for longer sequences of tasks. 
    It would happen, for example, when the tasks are orthogonal to each other.
    Their Theorem 6 fully characterizes when forgetting is zero for two tasks.
    
    \item When principal angles between task subspaces is large. We can think of the principal angles between task subspaces as distances between tasks. These principal angles are the same as the principal angles between null spaces, see Claim 19 of \citep{evron22a}. The following proposition says that if the smallest principal angle between the task null spaces is larger than $\pi/4$, then replay would not hurt forgetting. 
    \begin{proposition}\label{prop: far tasks no increase in forgetting}
        Suppose that assumptions \ref{assump: overparameterized lr}, \ref{assump: overparameterized lr},  \ref{assump: realizability} hold, and we are in the average case setup.
        Consider any two tasks where $\opnorm{\mP_2 \mP_1} \le \frac{\sqrt{2}}{2}$ and $\wstar \sim \normal (0, \id)$. Then forgetting with replay of any number of samples, averaged over $\wstar$ is never larger than forgetting without replay. That is, $\E_{\wstar} \brackets {F(\Tilde{\vw}_2)} \le \E_{\wstar} \brackets {F(\vw_2)}$.
    \end{proposition}
    Proof of this proposition is given in \cref{proof: close tasks proposition}.
    Note that in the two task setting, as described in \citet{evron22a}, forgetting would be small when the principal angles are away from $\pi/4$, so either larger or smaller but not intermediate values. Our results show that replay differentiates between these two scenarios.
    
\end{itemize}
This raises the interesting question of finding necessary and sufficient conditions under which replay would always be benign.

\section{Experiments} \label{sec:experiments}
We have included three sets of experiments. The first set of experiments explores the extent to which our theoretical results hold empirically when training with more complex (non-linear) networks.  
The other two experiments involve MNIST and are in a classification setting. In the second set of experiments, we replay one sample and examine how the class of the replayed sample affects forgetting. In the third set, we compare replay of different numbers of samples for two pairs of related task sequences.
Overall, the aim of these experiments is to verify the theoretical results and explore whether they can hold in more general settings, such as training with nonlinear models and classification.

\subsection{Empirical Evaluation of the Theoretical Results}
So far, we have shown that there are tasks that are realizable by linear models where sample replay increases forgetting when a linear model is trained sequentially. 
In this set of experiments, we verify these findings empirically and show that this behavior is not restricted to training with linear models. The experiments here show that there are (linear) tasks where replay can increase forgetting even when training nonlinear models on these tasks. 
We investigate the effect of replay on forgetting using a multi-layer perceptron (MLP) with one hidden layer and ReLU activations on a sequence of two tasks that are based on our worst case construction.

We consider two models: a linear model,
and a MLP with one hidden layer.
We consider two task sequence constructions.  The first construction is in $\reals^3$ and is based on the construction given in \cref{thm: avg case counter example}. The second task sequence is an extension of that construction into a higher dimensional space ($d=50$). Each sequence of tasks consists of a pair of tasks. The model is trained on the first task and then on the second task. Forgetting is then measured as the mean squared error of the final model on the first task's samples.
When training with replay,  $m$ samples from the first task are randomly selected without replacement and are combined with each batch during training on the second task. See  \cref{sec: experiment details} for further details on the experimental setup and how the construction in 
\cref{thm: avg case counter example} was extended to a higher dimensional input. 

\begin{figure}[t] 
\Centering
\begin{subfigure}{0.48\textwidth}
    \includegraphics[width=\linewidth]{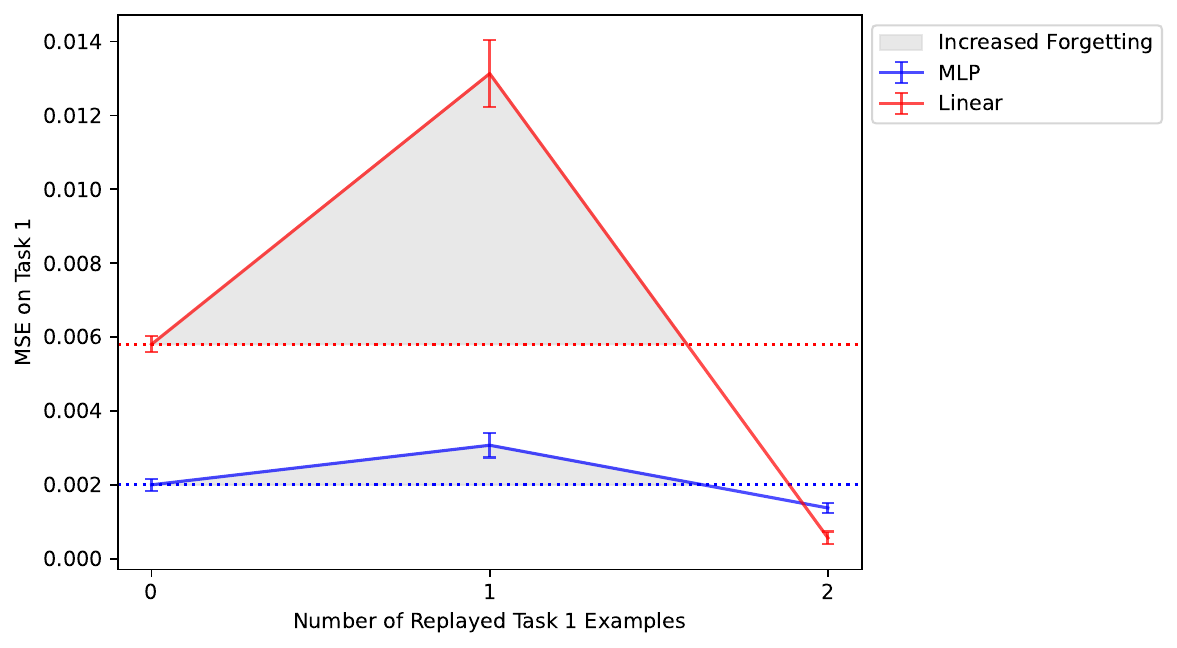}
    \caption{ The input data for the two tasks are given by the three dimensional construction given in \cref{thm: avg case counter example}. Each point is averaged over 150 runs. }
    \label{fig: replay errors 3 dim }
\end{subfigure}\hfill
\begin{subfigure} {0.48\textwidth}
    \includegraphics[width=\linewidth]{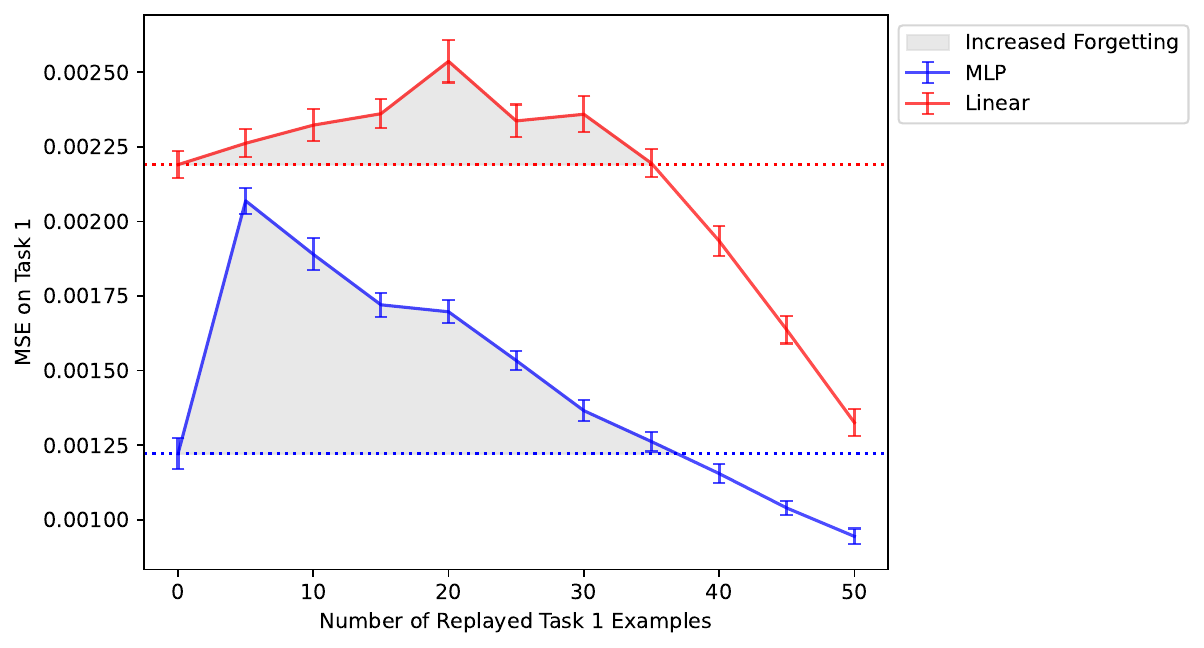}
     \caption{An extension of the three dimensional construction to $d=50$ displaying a similar behavior. Each point is averaged over 60 runs. 
    }
    \label{fig: replay errors 50 dim }
\end{subfigure}

\caption{Forgetting versus the number of replayed samples. Each plot shows forgetting of a linear model and a neural net with one hidden layer. We can see that forgetting initially increases with a small number of replay samples and then eventually decreases. The dashed lines show the baseline of no replay and the error bars indicate standard mean error. }
\label{fig: forgetting versus number of replayed samples}
\end{figure}

Another experiment included in \cref{app: angles and mlp}, provides empirical evidence that forgetting for the nonlinear models in this setting is affected by the angle between task null spaces through a mechanism similar to linear models. This provides some insight into the behavior of the nonlinear models on the continual linear regression problem, as seen in \cref{fig: forgetting versus number of replayed samples}.

\subsection{Experiments on MNIST}
The aim of the experiments in this section is to explore how the factors that led to harmful forgetting would affect replay in classification in more empirical settings.
In all the experiments in this section, a MLP with two hidden layers of size $256$ and ReLU activations was used. See \cref{appendix: experiments on MNIST} for more detail on these experiments.

\subsubsection{Rotated MNIST}

\begin{figure}[t] 
\Centering
\begin{subfigure}{0.48\textwidth}
    \includegraphics[width=\linewidth]{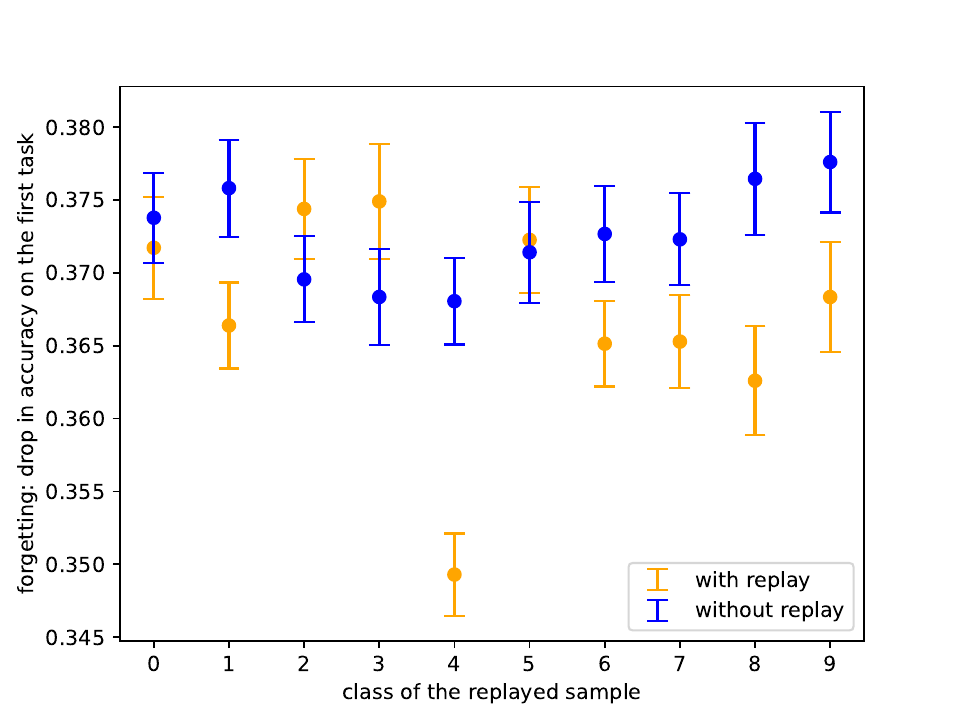}
    \caption{$45$ degrees rotation. Differences in means for classes $1$, $4$,  and $8$ are statistically significant. }
    \label{subfig: MNIST rotations 45}
\end{subfigure}\hfill
\begin{subfigure} {0.48\textwidth}
    \includegraphics[width=\linewidth]{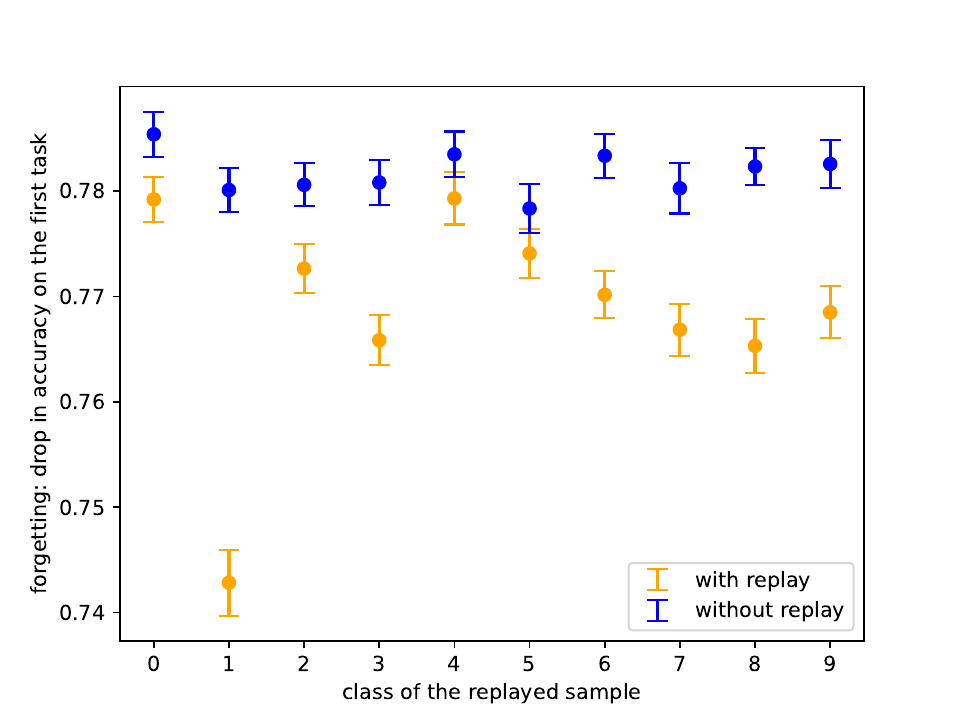}
     \caption{$90$ degrees rotation. Differences in the sample means are statistically significant except for $4$ and $5$.}
    \label{subfig: MNIST rotations 90}
\end{subfigure}
\caption{Class of the replayed sample affects forgetting. The x-axis shows the class of the replay sample while  the y-axis shows the amount of forgetting in Rotated MNIST. The points depict the averages and the error bars show mean standard error over $80$ runs. 
Comparing the average forgetting without replay to the one with replay of a single sample from each class shows that the effect of replay varies significantly across classes.
}
\label{fig: MNIST rotations}
\end{figure}

We study the role of replay samples in this experiment using Rotated MNIST \citep{GEM} in a task incremental setting.  We consider two tasks, where the first task is MNIST and the second task is Rotated MNIST, which has the same training data except that each digit
is rotated in the image. Forgetting is measured as the drop in classification accuracy in test data from the first task. 
In each run, after training a MLP on the first task, two copies of the network are made. One version  of the network is trained on the second task without replay and the other one is trained with replay of a single sample from the first task. The results of these experiments are shown in \cref{fig: MNIST rotations}.
On average, the extent to which replay decreases forgetting depends significantly on the class of the replayed sample. 
For example, we can see that for $45$ degrees rotation, replaying the digit $4$ reduces forgetting by about $3\%$, while replaying the digit $5$ does not seem to make much of a difference. 
Additionally, the relationship between tasks, in this case characterized by the degree of the rotation, also affects the behavior of replay with respect to the class of the replayed sample. 
For example, replaying a $4$ seems to be more beneficial in the  $45$ degrees rotation case than the $90$ degrees. 

\begin{figure}[t]
    \centering
    \includegraphics[width=0.5
\linewidth]{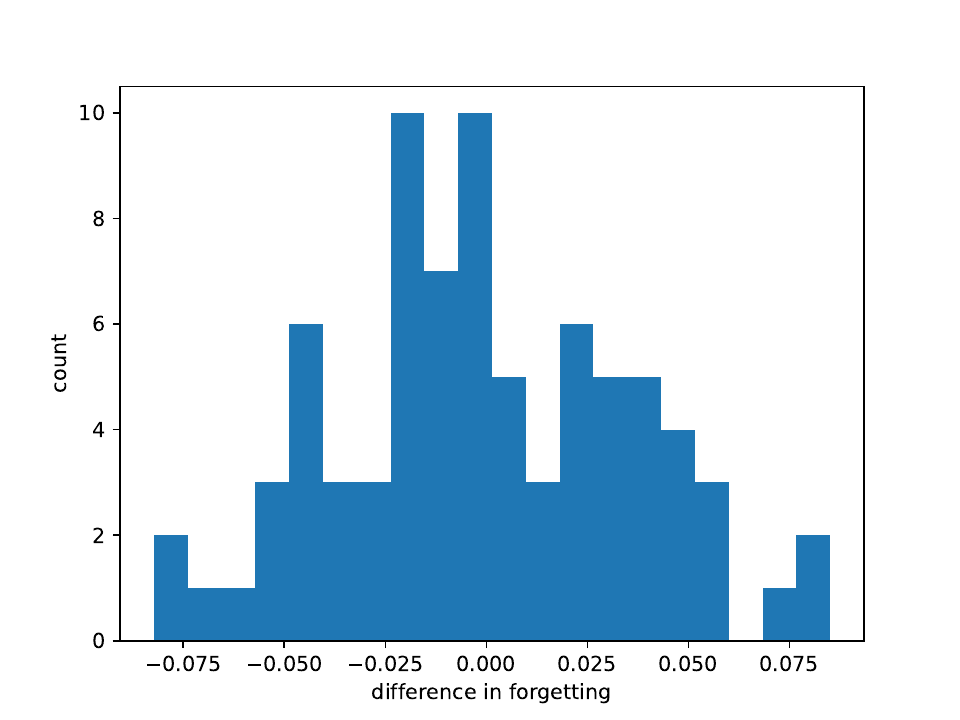}
    \caption{A histogram of differences in forgetting without replay and with replay of a digit $5$ sample in Rotated MNIST, where the second task is rotated by $45$ degrees.}
    \label{fig: forgetting diff hist}
\end{figure}

We take a closer look at replaying the digit $5$ in the $45$ degree rotation case, where replay seems to have minimal effect on forgetting. \cref{fig: forgetting diff hist} shows the distribution of differences in forgetting, that is, forgetting without replay minus forgetting with replay of a randomly selected digit $5$ sample.
We can see that even when on average this difference is not statistically significant, in many cases forgetting with replay exceeds without replay.
These results suggest that there could be significant differences in the efficacy of replay depending on which examples are replayed.

\subsubsection{Split MNIST}
\begin{figure}[t]
    \centering
    \includegraphics[width=0.5
\linewidth]{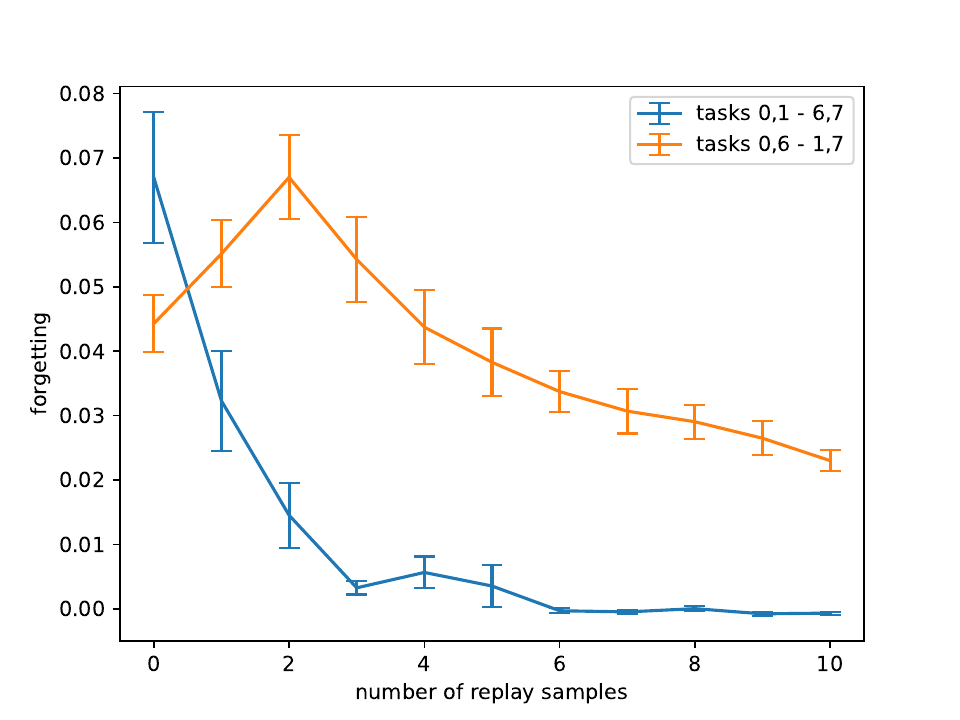}
    \caption{Role of tasks in replay for Split MNIST. For each task sequence, we have plotted the average forgetting against the number of replay samples. The error bars show mean standard error over $80$ runs. Values for $0$ replay samples show forgetting without replay.
    The differences in average forgetting across the two task sequences are statistically significant in all replay cases except for the no replay case. For the $0,6 - 1,7$ task sequence, the differences in the means for no replay and replay of $2$ samples is statistically significant, so the observed increase in forgetting is not noise. 
    }
    \label{fig: split MNIST}
\end{figure}

In this experiment, we study whether the relationship between tasks affects forgetting with replay in a class incremental setting. 
In one task sequence the first task involves discriminating $0$'s from $1$'s, and the second $6$'s from $7$'s; in the other sequence the first task is $0$ vs. $6$, the second $1$ vs. $7$.
\cref{fig: split MNIST} shows that in the first task sequence
replay consistently helps, but in the second forgetting initially increases, and only begins helping with 4 or more samples.
We hypothesize that the visual similarity of the digits in the first task in the $0,6$ - $1,7$ sequence makes the forgetting worse with a small number of replay samples, as the challenging discrimination task is sensitive to the selection of samples.

This empirical result bears important similarities to our theoretical results, in the average case setting.
The relationship between the tasks in a continual learning sequence determines the effect of replay on forgetting. And even in this more complicated, classification problem,  there exists sequences where forgetting can increase with replay.

\section{Related Work}
\label{sec:related}
For general background on continual or lifelong learning, see the surveys by \citet{SurveyDFC, PARISI201954,wang2024comprehensive}. 
Theoretical studies of continual learning, and especially replay have been relatively more scarce and recent.  
 Nevertheless, there are a few studies of continual learning in a linear setting.
\citet{doan21a} initiate the study of catastrophic forgetting of neural nets in the NTK regime, which also applies to linear models. 
\citet{goldfarb23a} study the effect of over-parameterization on forgetting in a linear regression setting for two tasks whose task subspaces are effectively low rank and picked randomly.
\citet{LJLS23} study generalization and a slightly different notion of forgetting in continual linear regression. They allow the tasks to be realized by different linear functions, while the task subspaces are essentially random.
\citet{li2023fixed} study the trade-off between stability and plasticity in a linear regression setting similar to \citep{evron22a}, with the distinction that they use $\ell_2$- regularization while training on the second task. 
\citet{shan2024order} study continual learning in deep learning from a statistical-mechanics point of view.

\citet{ICL} define a general continual learner that has no memory constraints and incurs zero forgetting. They propose an instantiation of it for continual linear regression that could require up to order $d^2$ bits of memory.
Additionally, through their framework, they derive uniform convergence type bounds and justify a form of sample replay where the learner keeps a task-balanced set of samples from previous tasks in memory and for each new task trains on the stored and new task's samples from scratch. Note that without knowing the exact constants in the bounds, this result does not provide much information on replay of fewer number of samples.
\citet{GDUMB} make a similar argument empirically and show that training from scratch on the samples stored in memory plus the most recent task's samples outperforms many methods that were specifically designed to address catastrophic forgetting.

There is a large body of empirical work on continual learning and methods used to mitigate forgetting, many of which use sample replay as a main strategy to address forgetting~\citep{rebuffi2017icarl,Chaudhry2019OnTE,wu2019large,aljundi2019online,caccia2021new}.
Although there has been some concern in the literature that replaying a small number of samples might lead to overfitting \citep{GEM}, \citet{Chaudhry2019OnTE} find that replay of even one sample per class improves performance and does not hurt generalization even when the model memorizes the replay samples. \citet{RehearsalRevealed} give a more complex picture of this in terms of the loss landscape. They find that while replay could keep the solution within a low loss region for previous tasks it could also pull the solution towards an unstable region, affecting generalization.
Motivated by forgetting in continual learning, \citet{toneva2018empirical} study forgetting of the examples across batches while training on a task.
They find that there are complex examples that are prone to forgetting across different model architectures.
Our results differ from these empirical results in that we focus on the existence of task sequences where replay would provably increase forgetting in a seemingly non adversarial setting.

In a recent work, \citet{BSR24} also study replay in continual linear regression. Their setup differs from ours in at least two significant ways.
First, the sequence of learned tasks lack any structured relationship, as samples for all the tasks are from the Gaussian $\normal(0, \id_d)$. In the over-parameterized regime where the number of samples per task is less than $d$, each task subspace is chosen uniformly at random.
Secondly, they allow the labels for each tasks to be generated by different linear functions, so their notion of task similarity is based on the distance between the parameters of each task rather than the relationship between subspaces. 
Assuming that all the tasks share the same linear function, their expression for sample replay is monotonically decreasing with respect to the number of samples.

\section{Conclusion}

We have shown that replay could increase forgetting in continual linear regression for some close pairs of tasks.
Our experimental results show that an increase in forgetting is not unique to linear models and can be  present when sequentially training with simple MLPs and even on a more natural class incremental continual learning problem.

This work opens a number of avenues for future study.
One interesting direction aims to form a method for identifying tasks in a dataset where replay could increase forgetting. 
A second direction considers relevant work in the cognitive science literature. For example, studies of Retrieval-Induced Forgetting \citep{RIF94, RIF00} find that retrieving some information in a category makes a subject more likely to forget the information in the same category that was not retrieved, which bears similarity to our finding
that replaying samples from a task could increase forgetting on the other samples in the same task that were not replayed.
Finally a valuable but challenging direction endeavors to develop a 
more complete characterization of when replay increases versus decreases forgetting.

\section{Acknowledgements}
This work is supported by the funds provided by the National Science Foundation and by DoD OUSD (R$\&$E) under Cooperative Agreement PHY-2229929 (The NSF AI Institute for Artificial and Natural Intelligence).

\bibliography{Collas25/cleaned_bibtex}

\begin{thebibliography}{40}
\providecommand{\natexlab}[1]{#1}
\providecommand{\url}[1]{\texttt{#1}}
\expandafter\ifx\csname urlstyle\endcsname\relax
  \providecommand{\doi}[1]{doi: #1}\else
  \providecommand{\doi}{doi: \begingroup \urlstyle{rm}\Url}\fi

\bibitem[Aljundi et~al.(2019{\natexlab{a}})Aljundi, Belilovsky, Tuytelaars, Charlin, Caccia, Lin, and Page-Caccia]{aljundi2019online}
Rahaf Aljundi, Eugene Belilovsky, Tinne Tuytelaars, Laurent Charlin, Massimo Caccia, Min Lin, and Lucas Page-Caccia.
\newblock Online continual learning with maximal interfered retrieval.
\newblock \emph{Advances in neural information processing systems}, 32, 2019{\natexlab{a}}.

\bibitem[Aljundi et~al.(2019{\natexlab{b}})Aljundi, Lin, Goujaud, and Bengio]{aljundi2019gradient}
Rahaf Aljundi, Min Lin, Baptiste Goujaud, and Yoshua Bengio.
\newblock Gradient based sample selection for online continual learning.
\newblock \emph{Advances in neural information processing systems}, 32, 2019{\natexlab{b}}.

\bibitem[Anderson et~al.(1994)Anderson, Bjork, and Bjork]{RIF94}
M.~C. Anderson, R.~A. Bjork, and E.~L. Bjork.
\newblock Remembering can cause forgetting: Retrieval dynamics in long-term memory.
\newblock \emph{Journal of Experimental Psychology: Learning, Memory, and Cognition}, \penalty0 (20(5)):\penalty0 1063--1087, 1994.

\bibitem[Anderson et~al.(2000)Anderson, Bjork, and Bjork]{RIF00}
M~C Anderson, E~L Bjork, and R~A Bjork.
\newblock Retrieval-induced forgetting: evidence for a recall-specific mechanism.
\newblock \emph{Journal of Experimental Psychology: Learning, Memory, and Cognition}, 7\penalty0 (3):\penalty0 522--530, Sep 2000.

\bibitem[Banayeeanzade et~al.(2024)Banayeeanzade, Soltanolkotabi, and Rostami]{BSR24}
Mohammadamin Banayeeanzade, Mahdi Soltanolkotabi, and Mohammad Rostami.
\newblock Theoretical insights into overparameterized models in multi-task and replay-based continual learning, 2024.

\bibitem[Caccia et~al.(2021)Caccia, Aljundi, Asadi, Tuytelaars, Pineau, and Belilovsky]{caccia2021new}
Lucas Caccia, Rahaf Aljundi, Nader Asadi, Tinne Tuytelaars, Joelle Pineau, and Eugene Belilovsky.
\newblock New insights on reducing abrupt representation change in online continual learning.
\newblock \emph{arXiv preprint arXiv:2104.05025}, 2021.

\bibitem[Chaudhry et~al.(2019)Chaudhry, Rohrbach, Elhoseiny, Ajanthan, Dokania, Torr, and Ranzato]{Chaudhry2019OnTE}
Arslan Chaudhry, Marcus Rohrbach, Mohamed Elhoseiny, Thalaiyasingam Ajanthan, Puneet~Kumar Dokania, Philip H.~S. Torr, and Marc'Aurelio Ranzato.
\newblock On tiny episodic memories in continual learning.
\newblock \emph{arXiv: Learning}, 2019.

\bibitem[Chen et~al.(2022)Chen, Papadimitriou, and Peng]{Chen2022MemoryBF}
Xi~Chen, Christos Papadimitriou, and Binghui Peng.
\newblock Memory bounds for continual learning.
\newblock \emph{2022 IEEE 63rd Annual Symposium on Foundations of Computer Science (FOCS)}, pp.\  519--530, 2022.

\bibitem[Dasgupta \& Gupta(2003)Dasgupta and Gupta]{Dasgupta2003AnEP}
Sanjoy Dasgupta and Anupam Gupta.
\newblock An elementary proof of a theorem of johnson and lindenstrauss.
\newblock \emph{Random Structures \& Algorithms}, 22, 2003.

\bibitem[De~Lange et~al.(2022)De~Lange, Aljundi, Masana, Parisot, Jia, Leonardis, Slabaugh, and Tuytelaars]{SurveyDFC}
Matthias De~Lange, Rahaf Aljundi, Marc Masana, Sarah Parisot, Xu~Jia, Aleš Leonardis, Gregory Slabaugh, and Tinne Tuytelaars.
\newblock A continual learning survey: Defying forgetting in classification tasks.
\newblock \emph{IEEE Transactions on Pattern Analysis and Machine Intelligence}, 44\penalty0 (7):\penalty0 3366--3385, 2022.
\newblock \doi{10.1109/TPAMI.2021.3057446}.

\bibitem[Doan et~al.(2021)Doan, Abbana~Bennani, Mazoure, Rabusseau, and Alquier]{doan21a}
Thang Doan, Mehdi Abbana~Bennani, Bogdan Mazoure, Guillaume Rabusseau, and Pierre Alquier.
\newblock A theoretical analysis of catastrophic forgetting through the ntk overlap matrix.
\newblock In Arindam Banerjee and Kenji Fukumizu (eds.), \emph{Proceedings of The 24th International Conference on Artificial Intelligence and Statistics}, volume 130 of \emph{Proceedings of Machine Learning Research}, pp.\  1072--1080. PMLR, 13--15 Apr 2021.

\bibitem[Evron et~al.(2022)Evron, Moroshko, Ward, Srebro, and Soudry]{evron22a}
Itay Evron, Edward Moroshko, Rachel Ward, Nathan Srebro, and Daniel Soudry.
\newblock How catastrophic can catastrophic forgetting be in linear regression?
\newblock In Po-Ling Loh and Maxim Raginsky (eds.), \emph{Proceedings of Thirty Fifth Conference on Learning Theory}, volume 178 of \emph{Proceedings of Machine Learning Research}, pp.\  4028--4079. PMLR, 02--05 Jul 2022.

\bibitem[Evron et~al.(2023)Evron, Moroshko, Buzaglo, Khriesh, Marjieh, Srebro, and Soudry]{evron2023classification}
Itay Evron, Edward Moroshko, Gon Buzaglo, Maroun Khriesh, Badea Marjieh, Nathan Srebro, and Daniel Soudry.
\newblock Continual learning in linear classification on separable data.
\newblock In \emph{Proceedings of the 40th International Conference on Machine Learning}, ICML'23. JMLR.org, 2023.

\bibitem[Goldfarb \& Hand(2023)Goldfarb and Hand]{goldfarb23a}
Daniel Goldfarb and Paul Hand.
\newblock Analysis of catastrophic forgetting for random orthogonal transformation tasks in the overparameterized regime.
\newblock In Francisco Ruiz, Jennifer Dy, and Jan-Willem van~de Meent (eds.), \emph{Proceedings of The 26th International Conference on Artificial Intelligence and Statistics}, volume 206 of \emph{Proceedings of Machine Learning Research}, pp.\  2975--2993. PMLR, 25--27 Apr 2023.

\bibitem[Gunasekar et~al.(2018)Gunasekar, Lee, Soudry, and Srebro]{gunasekar18a}
Suriya Gunasekar, Jason Lee, Daniel Soudry, and Nathan Srebro.
\newblock Characterizing implicit bias in terms of optimization geometry.
\newblock In Jennifer Dy and Andreas Krause (eds.), \emph{Proceedings of the 35th International Conference on Machine Learning}, volume~80 of \emph{Proceedings of Machine Learning Research}, pp.\  1832--1841. PMLR, 10--15 Jul 2018.

\bibitem[Horn \& Johnson(1985)Horn and Johnson]{Horn_Johnson_1985}
Roger~A. Horn and Charles~R. Johnson.
\newblock \emph{Matrix Analysis}.
\newblock Cambridge University Press, 1985.

\bibitem[Kingma \& Ba(2014)Kingma and Ba]{Kingma2014AdamAM}
Diederik~P. Kingma and Jimmy Ba.
\newblock Adam: A method for stochastic optimization.
\newblock \emph{CoRR}, abs/1412.6980, 2014.

\bibitem[Kirkpatrick et~al.(2017)Kirkpatrick, Pascanu, Rabinowitz, Veness, Desjardins, Rusu, Milan, Quan, Ramalho, Grabska-Barwinska, et~al.]{kirkpatrick2017overcoming}
James Kirkpatrick, Razvan Pascanu, Neil Rabinowitz, Joel Veness, Guillaume Desjardins, Andrei~A Rusu, Kieran Milan, John Quan, Tiago Ramalho, Agnieszka Grabska-Barwinska, et~al.
\newblock Overcoming catastrophic forgetting in neural networks.
\newblock \emph{Proceedings of the national academy of sciences}, 114\penalty0 (13):\penalty0 3521--3526, 2017.

\bibitem[Li et~al.(2023)Li, Wu, and Braverman]{li2023fixed}
Haoran Li, Jingfeng Wu, and Vladimir Braverman.
\newblock Fixed design analysis of regularization-based continual learning.
\newblock In \emph{Proceedings of the 2nd Conference on Lifelong Learning Agents (CoLLAs)}, pp.\  513--533, 2023.
\newblock arXiv:2303.10263.

\bibitem[Li \& Hoiem(2017)Li and Hoiem]{li2017learning}
Zhizhong Li and Derek Hoiem.
\newblock Learning without forgetting.
\newblock \emph{IEEE transactions on pattern analysis and machine intelligence}, 40\penalty0 (12):\penalty0 2935--2947, 2017.

\bibitem[Lin(1992)]{lin1992reinforcement}
Long-Ji Lin.
\newblock \emph{Reinforcement learning for robots using neural networks}.
\newblock Carnegie Mellon University, 1992.

\bibitem[Lin et~al.(2023)Lin, Ju, Liang, and Shroff]{LJLS23}
Sen Lin, Peizhong Ju, Yingbin Liang, and Ness Shroff.
\newblock Theory on forgetting and generalization of continual learning.
\newblock In \emph{Proceedings of the 40th International Conference on Machine Learning}, ICML'23. JMLR.org, 2023.

\bibitem[Lopez-Paz \& Ranzato(2017)Lopez-Paz and Ranzato]{GEM}
David Lopez-Paz and Marc'Aurelio Ranzato.
\newblock Gradient episodic memory for continual learning.
\newblock In \emph{Proceedings of the 31st International Conference on Neural Information Processing Systems}, NIPS'17, pp.\  6470–6479, Red Hook, NY, USA, 2017. Curran Associates Inc.
\newblock ISBN 9781510860964.

\bibitem[McCloskey \& Cohen(1989)McCloskey and Cohen]{mccloskey1989catastrophic}
Michael McCloskey and Neal~J Cohen.
\newblock Catastrophic interference in connectionist networks: The sequential learning problem.
\newblock In \emph{Psychology of learning and motivation}, volume~24, pp.\  109--165. Elsevier, 1989.

\bibitem[Oudiette \& Paller(2013)Oudiette and Paller]{Oudiette2013}
Delphine Oudiette and Ken Paller.
\newblock Upgrading the sleeping brain with targeted memory reactivation.
\newblock \emph{Trends in Cognitive Science}, 3\penalty0 (17):\penalty0 pp. 142--149, 2013.

\bibitem[Parisi et~al.(2019)Parisi, Kemker, Part, Kanan, and Wermter]{PARISI201954}
German~I. Parisi, Ronald Kemker, Jose~L. Part, Christopher Kanan, and Stefan Wermter.
\newblock Continual lifelong learning with neural networks: A review.
\newblock \emph{Neural Networks}, 113:\penalty0 54--71, 2019.
\newblock ISSN 0893-6080.
\newblock \doi{https://doi.org/10.1016/j.neunet.2019.01.012}.

\bibitem[Peng et~al.(2023)Peng, Giampouras, and Vidal]{ICL}
Liangzu Peng, Paris~V. Giampouras, and Ren\'{e} Vidal.
\newblock The ideal continual learner: an agent that never forgets.
\newblock In \emph{Proceedings of the 40th International Conference on Machine Learning}, ICML'23. JMLR.org, 2023.

\bibitem[Prabhu et~al.(2020)Prabhu, Torr, and Dokania]{GDUMB}
Ameya Prabhu, Philip H.~S. Torr, and Puneet~K. Dokania.
\newblock Gdumb: A simple approach that questions our progress in continual learning.
\newblock In Andrea Vedaldi, Horst Bischof, Thomas Brox, and Jan-Michael Frahm (eds.), \emph{Computer Vision -- ECCV 2020}, pp.\  524--540, Cham, 2020. Springer International Publishing.
\newblock ISBN 978-3-030-58536-5.

\bibitem[Rasch \& Born(2007)Rasch and Born]{RaschBorn}
Bjorn Rasch and Jan Born.
\newblock Maintaining memories by reactivation.
\newblock \emph{Current Opinions in Neurobiology}, 6\penalty0 (17):\penalty0 pp. 698--703, 2007.

\bibitem[Rebuffi et~al.(2017)Rebuffi, Kolesnikov, Sperl, and Lampert]{rebuffi2017icarl}
Sylvestre-Alvise Rebuffi, Alexander Kolesnikov, Georg Sperl, and Christoph~H Lampert.
\newblock icarl: Incremental classifier and representation learning.
\newblock In \emph{Proceedings of the IEEE conference on Computer Vision and Pattern Recognition}, pp.\  2001--2010, 2017.

\bibitem[Rolnick et~al.(2019)Rolnick, Ahuja, Schwarz, Lillicrap, and Wayne]{rolnick2019experience}
David Rolnick, Arun Ahuja, Jonathan Schwarz, Timothy Lillicrap, and Gregory Wayne.
\newblock Experience replay for continual learning.
\newblock \emph{Advances in Neural Information Processing Systems}, 32, 2019.

\bibitem[Shan et~al.(2024)Shan, Li, and Sompolinsky]{shan2024order}
Haozhe Shan, Qianyi Li, and Haim Sompolinsky.
\newblock Order parameters and phase transitions of continual learning in deep neural networks.
\newblock \emph{arXiv preprint arXiv:2407.10315}, 2024.

\bibitem[Tiwari et~al.(2022)Tiwari, Killamsetty, Iyer, and Shenoy]{tiwari2022gcr}
Rishabh Tiwari, Krishnateja Killamsetty, Rishabh Iyer, and Pradeep Shenoy.
\newblock Gcr: Gradient coreset based replay buffer selection for continual learning.
\newblock In \emph{Proceedings of the IEEE/CVF Conference on Computer Vision and Pattern Recognition}, pp.\  99--108, 2022.

\bibitem[Toneva et~al.(2018)Toneva, Sordoni, Combes, Trischler, Bengio, and Gordon]{toneva2018empirical}
Mariya Toneva, Alessandro Sordoni, Remi Tachet~des Combes, Adam Trischler, Yoshua Bengio, and Geoffrey~J Gordon.
\newblock An empirical study of example forgetting during deep neural network learning.
\newblock \emph{arXiv preprint arXiv:1812.05159}, 2018.

\bibitem[van~de Ven et~al.(2020)van~de Ven, Siegelmann, and Tolias]{BrainGR}
Gido~M. van~de Ven, Hava~T. Siegelmann, and Andreas~S. Tolias.
\newblock Brain-inspired replay for continual learning with artificial neural networks.
\newblock \emph{Nature Communications}, 11\penalty0 (1):\penalty0 4069, 2020.
\newblock \doi{10.1038/s41467-020-17866-2}.

\bibitem[Verwimp et~al.(2021)Verwimp, De~Lange, and Tuytelaars]{RehearsalRevealed}
Eli Verwimp, Matthias De~Lange, and Tinne Tuytelaars.
\newblock Rehearsal revealed: The limits and merits of revisiting samples in continual learning.
\newblock In \emph{2021 IEEE/CVF International Conference on Computer Vision (ICCV)}, pp.\  9365--9374, 2021.
\newblock \doi{10.1109/ICCV48922.2021.00925}.

\bibitem[Wang et~al.(2024)Wang, Zhang, Su, and Zhu]{wang2024comprehensive}
Liyuan Wang, Xingxing Zhang, Hang Su, and Jun Zhu.
\newblock A comprehensive survey of continual learning: Theory, method and application, 2024.

\bibitem[Wu et~al.(2019)Wu, Chen, Wang, Ye, Liu, Guo, and Fu]{wu2019large}
Yue Wu, Yinpeng Chen, Lijuan Wang, Yuancheng Ye, Zicheng Liu, Yandong Guo, and Yun Fu.
\newblock Large scale incremental learning.
\newblock In \emph{Proceedings of the IEEE/CVF conference on computer vision and pattern recognition}, pp.\  374--382, 2019.

\bibitem[Zenke et~al.(2017)Zenke, Poole, and Ganguli]{zenke2017continual}
Friedemann Zenke, Ben Poole, and Surya Ganguli.
\newblock Continual learning through synaptic intelligence.
\newblock In \emph{International conference on machine learning}, pp.\  3987--3995. PMLR, 2017.

\bibitem[Zhang et~al.(2021)Zhang, Bengio, Hardt, Recht, and Vinyals]{rethinking_generalization}
Chiyuan Zhang, Samy Bengio, Moritz Hardt, Benjamin Recht, and Oriol Vinyals.
\newblock Understanding deep learning (still) requires rethinking generalization.
\newblock \emph{Commun. ACM}, 64\penalty0 (3):\penalty0 107–115, feb 2021.
\newblock ISSN 0001-0782.
\newblock \doi{10.1145/3446776}.

\end{thebibliography}
\bibliographystyle{collas2025_conference}

\appendix
\newpage

\section{Additional Background and Definitions}\label{sec: additional background}

\subsection{Derivation of \cref{eq: residual error}}
\label{sec: derivation of recursive error}
The following is based on \citet{evron22a} and included here for completeness.
More information on the properties of this solution and different forms of it are provided in their work.

We start with the closed form solution of \cref{eq: learning procedure}. 
Recall that $\mPi_t, \mP_t$ are orthogonal projections onto the spans of  row spaces and null space of $\mX_t$ respectively.
Every solution to the equation 
$\mX_t \vw = \vy_t$ can be written as 
$ \vw = \mPi_t \vw^* + \mP_t \vv$ for some vector $\vv$. It is then easy to see that $\mP_t w_{t-1}$ would minimize $\twonorm{\vw - \vw_{t-1}}$ and is in the null space of $\mX_t$, so the closed form solution would be $\vw_t = \mPi_t \vw^* + \mP_t \vw_{t-1}$.
Subtracting $\vw^*$ from both sides would give \cref{eq: residual error}.

\subsection{Derivation of \cref{eq: forgetting sequential projections}} \label{sec: derivation of basic forgetting}
This derivation is also in given in \citet{evron22a} and is included here for completeness. 
Using \cref{assump: realizability}, we can write each  $\vy_t = \mX_t \vw^*$ and the forgetting as 
\begin{align} \label{eq: forgetting w star}
    F_S(\vw_T) = \frac{1}{T-1} \sum_{t = 1}^{T-1} \twonorm{\mX_t (\vw_T - \vw^*)}^2.
\end{align}

Applying \cref{eq: residual error} repeatedly, the parameter error vector after task $T$  would be
\begin{align}
    \vw_T - \vw^* = & \mP_T (\vw_{T-1} - \vw^*) 
    = 
    \mP_T \dots \mP_1 (\vw_0 - \vw^*) = - \mP_T \dots \mP_1 \vw^*.
\end{align}

Plugging the equation above into each term in forgetting, we have
\begin{align}
    F_S(\vw_T) = & \frac{1}{T-1} \sum_{t = 1}^{T-1} \twonorm{\mX_t (\vw_T - \vw^*)}^2
     =  \frac{1}{T-1} \sum_{t = 1}^{T-1} \twonorm{\mX_t \mP_T \mP_{T-1} \dots \mP_1 \vw^*}^2.
\end{align}

\subsection{The Angle between Null Spaces and Forgetting} \label{sec: explanation for angle between null spaces}
Principal angles between two subspaces are a generalization of angles between two vectors. 
We start with the following simple two task example. Let $\vw^*$ be an arbitrary unit vector and consider two tasks whose null spaces are $\mP_1 = \bfa_1 \bfa_1^\top$ and $\mP_2 = \bfa_2 \bfa_2^\top$, where $\bfa_1, \bfa_2$ are unit vectors such that $\twonorm{\bfa_1^\top \vw^*} > 0$.  
In this two task case, the expected forgetting, given in \cref{eq: expected forgetting},  is proportional to  $ \twonorm{\mPi_1 \mP_2 \mP_1 \vw^*}^2$.

\begin{figure}[h]

    \centering

    \includegraphics[width=0.3\textwidth]{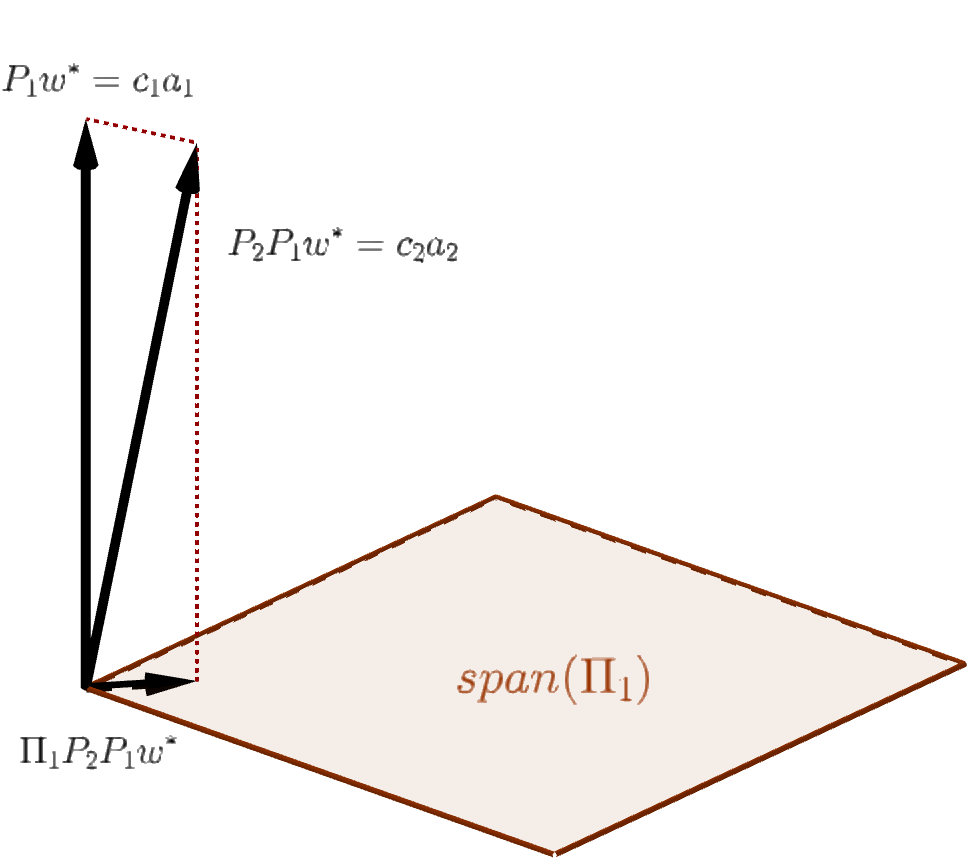} \hfill
    \includegraphics[width=0.3\textwidth]{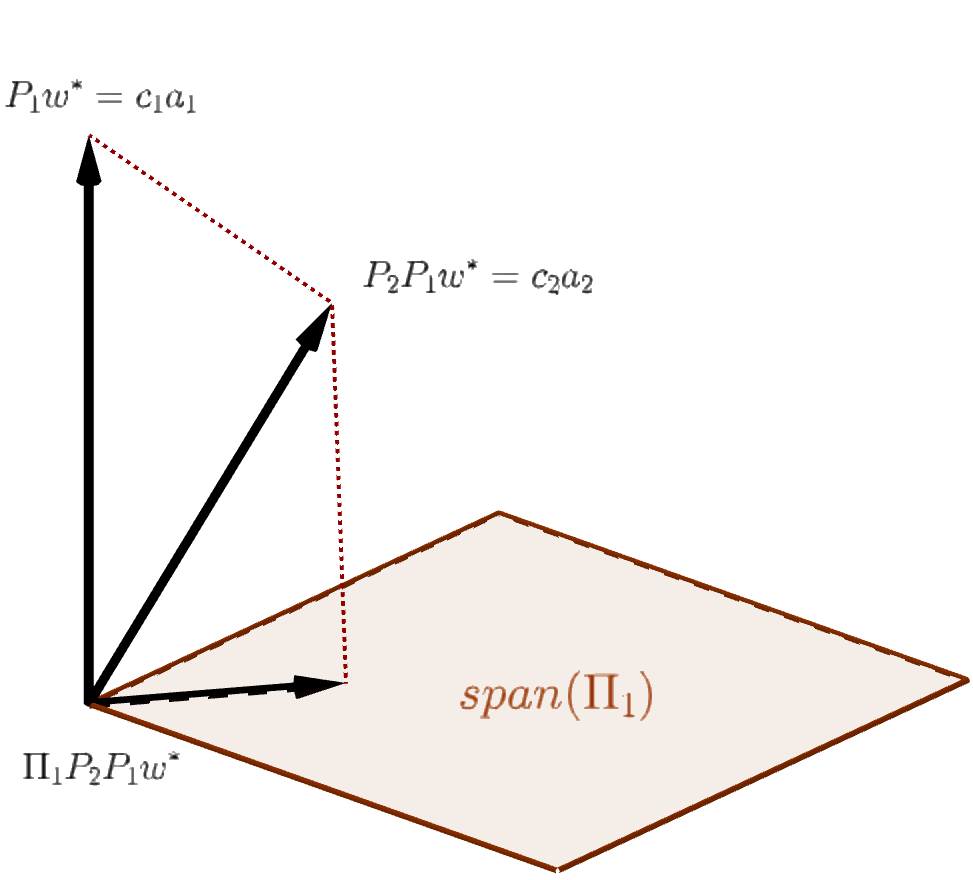}\hfill
    \includegraphics[width=0.3\textwidth]{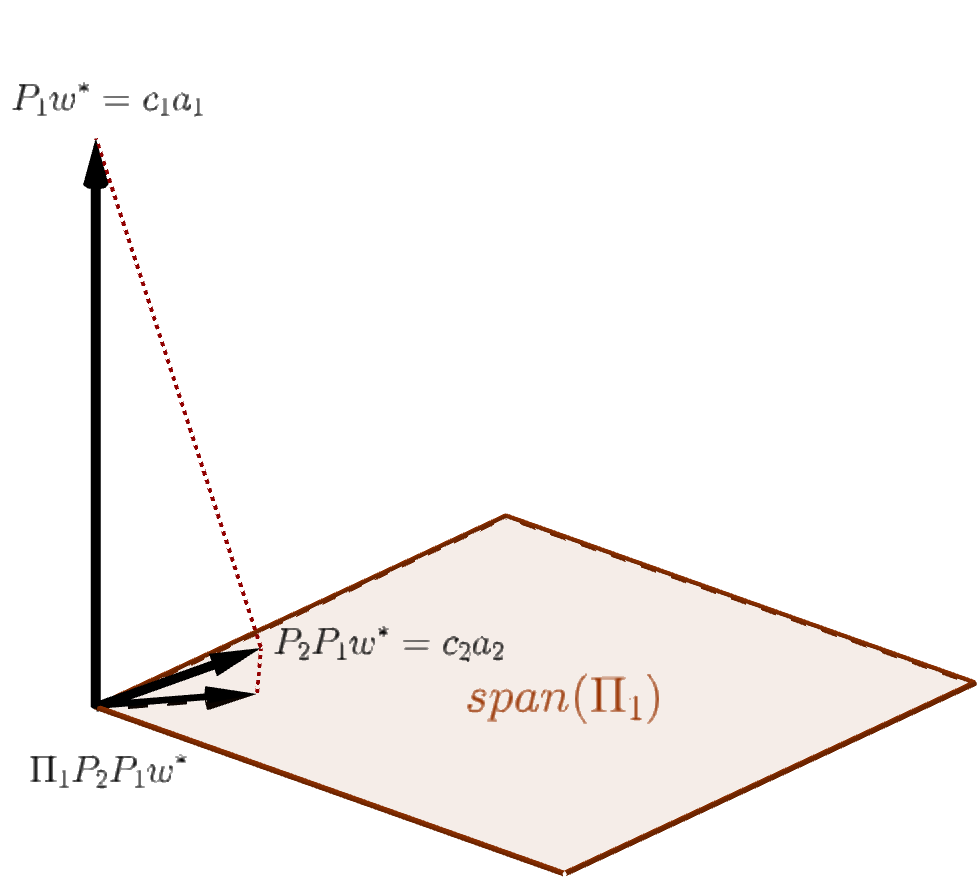}

    \caption{A simple example to demonstrate how the angle between tasks with one dimensional null spaces affects forgetting. $\bfa_1, \bfa_2$ span the null spaces of the first and second tasks respectively.
    The two vectors that are outside $\sp{\mPi_1}$ show the projections $\mP_1 \vw^*$ and $\mP_2 \mP_1 \vw^*$ which will be in the directions of $\bfa_1$ and $\bfa_2$ as marked. 
    Each display shows the effect of the angle between the null spaces, which is the angle between $\bfa_1$ and $\bfa_2$, on forgetting, which would be $\twonorm{\mPi_1 \mP_2 \mP_1 \vw^*}$. As the angle between $\bfa_1$ and $\bfa_2$ increases (from the left to right display), the forgetting first increases and then decreases. Forgetting is maximized when the angle between $\bfa_1$ and $\bfa_2$ is $\pi/4$.}
    \label{fig: angles and forgetting}
\end{figure}

Note that $\mP_1 \vw^* = \bfa_1 \bfa_1^\top \vw^*=  c_1 \bfa_1$ and $\mP_2 \mP_1 \vw^* = c_1 \bfa_2 \bfa_2^\top \bfa_1= c_2 \bfa_2$ with $c_1 = \bfa_1^\top \vw^*$, and $c_2 = c_1 \bfa_2^\top \bfa_1$.
\Cref{fig: angles and forgetting} in \cref{sec: explanation for angle between null spaces} shows how the norm of $\mPi_1 \mP_2 \mP_1 \vw^*$ depends on the angle between $\bfa_1$ and $\bfa_2$. When the angle between  $\bfa_1$ and $\bfa_2$ is small, as in the left display, forgetting will be small. 
In fact if the angle was zero, then forgetting would be zero. At the other extreme is when the tasks are almost orthogonal, as shown in the right display, which would also result in less forgetting. 
\citet{evron22a} show that the maximum forgetting (over the choice of $\vw^*$ and $\bfa_2$) in this two task setting is proportional to $(\bfa_1^\top \bfa_2)^2 \paren{ 1 - (\bfa_1^\top \bfa_2)^2}$ which is maximized when the angle between $\bfa_1$ and $\bfa_2$ is $\pi/4$.

\section{Intuition for the proofs}
\subsection{Intuition for the worst case result}
\label{appendix: worst case intuition}
\begin{figure} [H]
    \centering
    \caption{Replay can transfer error between samples. The first task consists of two samples $\vx_1$ and $\vx_2$, while the second task has one sample $\vx_3$. All of the plots display the three samples $\vx_1, \vx_2, \vx_3$, the target parameter vector $\vw^*$, and the iterates without replay $\vw_1$ and $\vw_2$ from different angles. The vector $v_2$, which is orthogonal to the samples $\vx_1$ and $\vx_3$, is also displayed to show the orientation of the plots. The left figure in each row shows the general position of vectors of interest.  Rest of the figures in the first row focus on the error of the final iterate $\vw_2$ on the first task's samples without replay.
    The second row, additionally, shows the final iterate $\Tilde{\vw}_2$ after replay of $\vx_2$, and the projection of $\Tilde{\vw}_2$ onto the first task's samples. In each case, intersection of the dashed line originating from parameter vectors $\vw^*, \vw_1, \dots$ with the sample $\vx_2$ or $\vx_1$ shows the projection of the parameter vector onto that sample. 
    The error of each iterate $\vw_2$ and $\Tilde{\vw}_2$ along a sample is marked in red by the discrepancy between its' projection and the projection of $\vw^*$ onto the sample. }
    \begin{subfigure}{\textwidth}
    \includegraphics[width=0.30\textwidth]{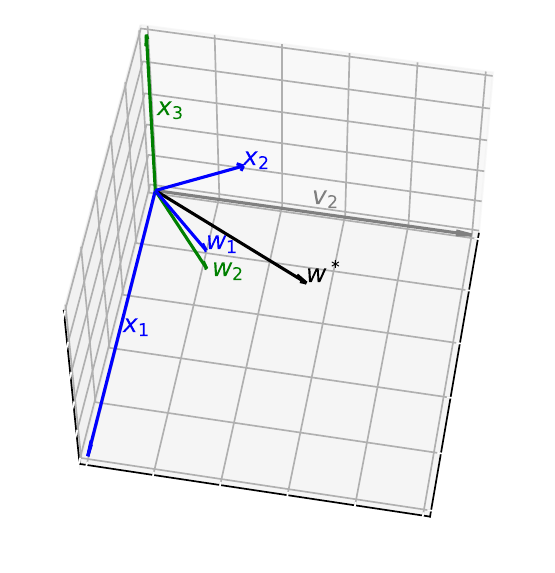} \hfill
    \includegraphics[width=0.34\textwidth]{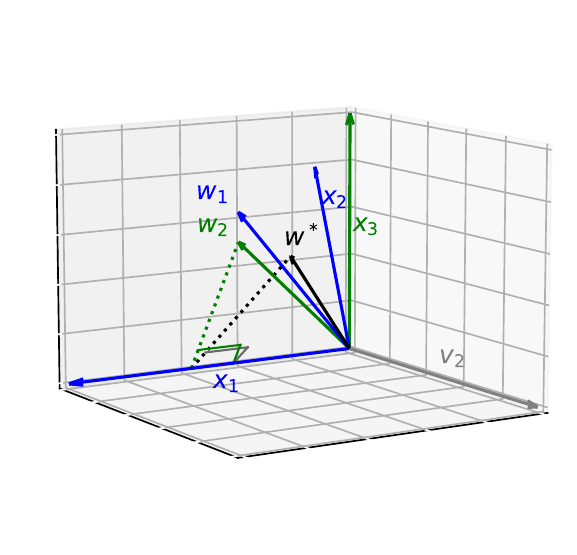} \hfill
    \includegraphics[width=0.345\textwidth]{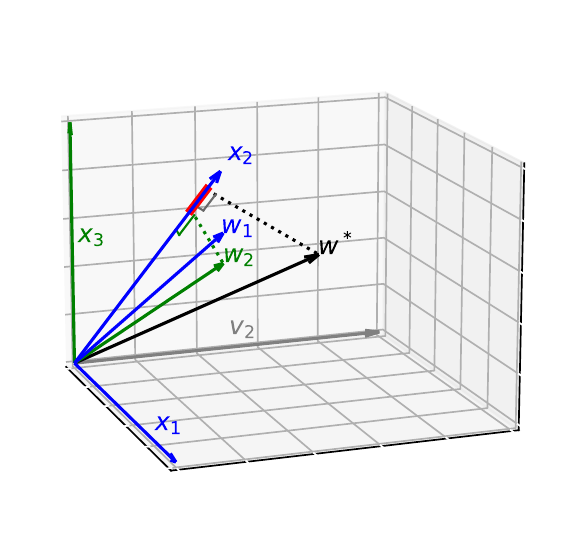}
    \caption{Without replay: The first iterate $\vw_1$ is in the span of $\vx_1$ and $\vx_2$.  The final iterate $\vw_2$ is obtained by training on $(\vx_3, \vx_3^\top \vw^*)$ starting from $\vw_1$.  The change from $\vw_1$ to $\vw_2$ is in the direction of $-\vx_3$, and since $\vx_3$ is orthogonal to $\vx_1$, $\vw_2$ has no error along $\vx_1$. We can see this in the center figure where projections of $\vw_2$ and $\vw^*$ onto $\vx_1$ coincide. 
    As we can see in the right figure, $\vw_2$ will have some error along $\vx_2$, since $\vx_3$ is not orthogonal to $\vx_2$. The discrepancy in the projections of $\vw_2$ and $\vw^*$ onto $\vx_2$ is shown in red. }
     \label{fig: worst case without replay}
    \end{subfigure}\hfill

    \begin{subfigure}{\textwidth}
        \includegraphics[width=0.32\textwidth]{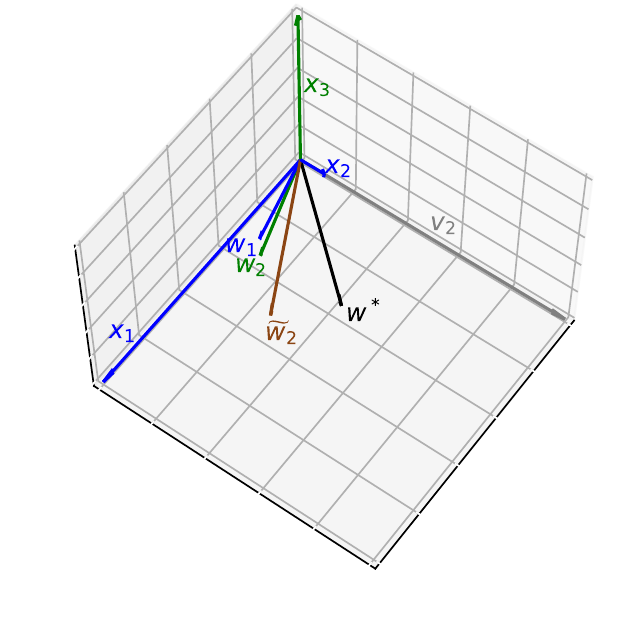}\hfill
        \includegraphics[width=0.32\textwidth]{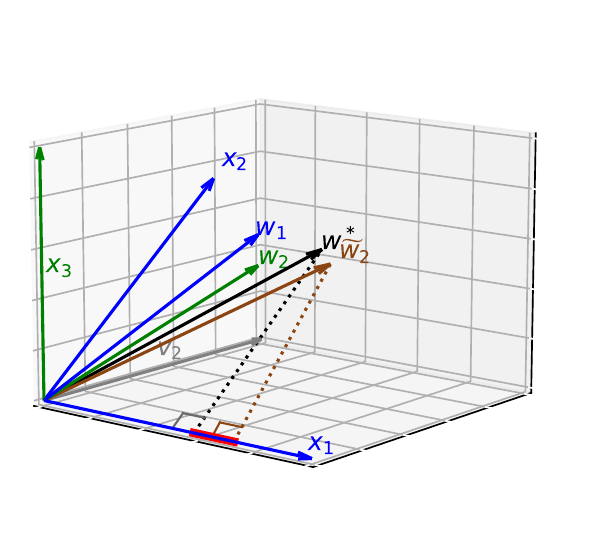}\hfill
        \includegraphics[width=0.32\textwidth]{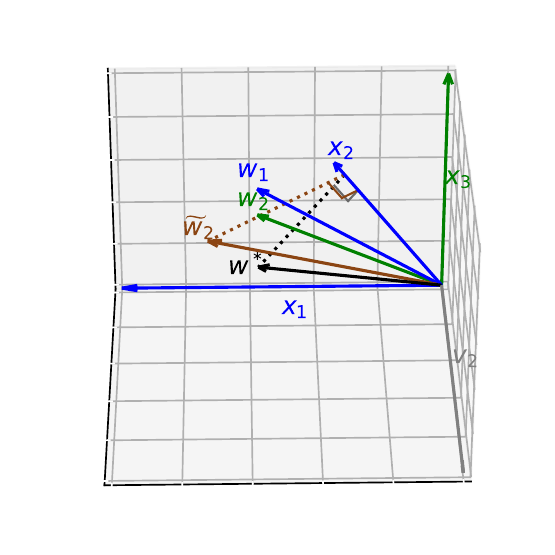}
        \caption{With replay: Since $\vw_2$ had error on $\vx_2$, we consider replaying $\vx_2$. The first display shows the iterate after replay of $\vx_2$, $\Tilde{\vw}_2$ relative to the other iterates and examples.  Since $\Tilde{\vw}_2$ changes in the direction of $\vx_2$ in addition to changing in the direction of $- \vx_3$, it also moves in the direction of $\vx_2$. This causes it to incur some error in the direction of $\vx_1$ as seen in the center figure.  The right figure shows that in contrast to $\vw_2$, $\Tilde{\vw}_2$ does not have any error along $\vx_2$, since $\vx_2$ was just replayed. 
        }
        \label{fig: worst case with replay}
    \end{subfigure}
    
    \label{fig: worst case intuition }
\end{figure}

\subsection{Intuition for the lower dimensional average case result}
\begin{figure} 
    \centering
    \begin{subfigure}{0.25\textwidth}
      \includegraphics[width=\linewidth]{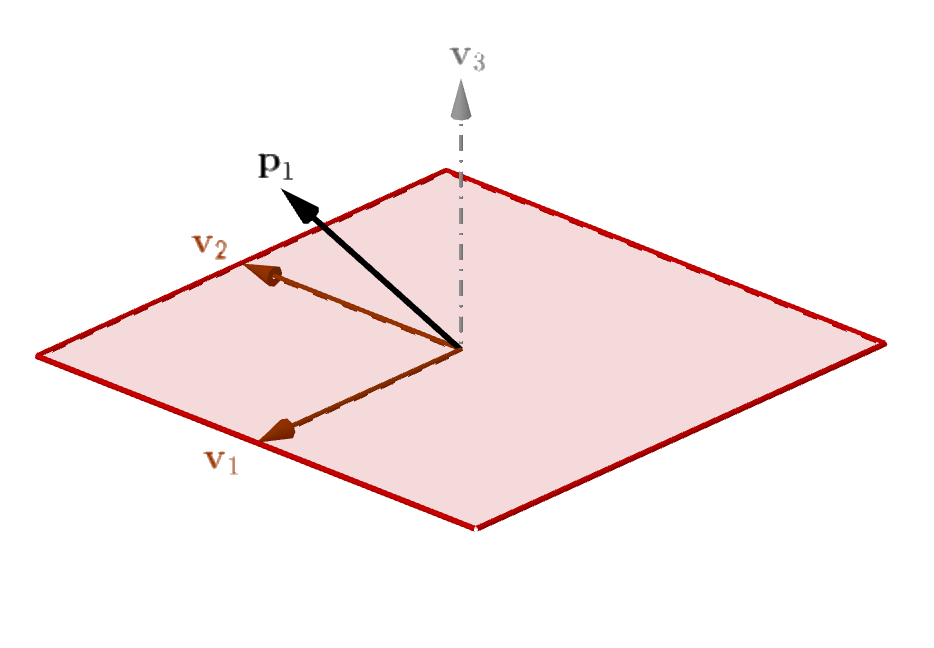}
        \caption{}
        \label{fig: avg case null spaces}
    \end{subfigure}\hfill
    \begin{subfigure}{0.25\textwidth}
        \includegraphics[width=\linewidth]{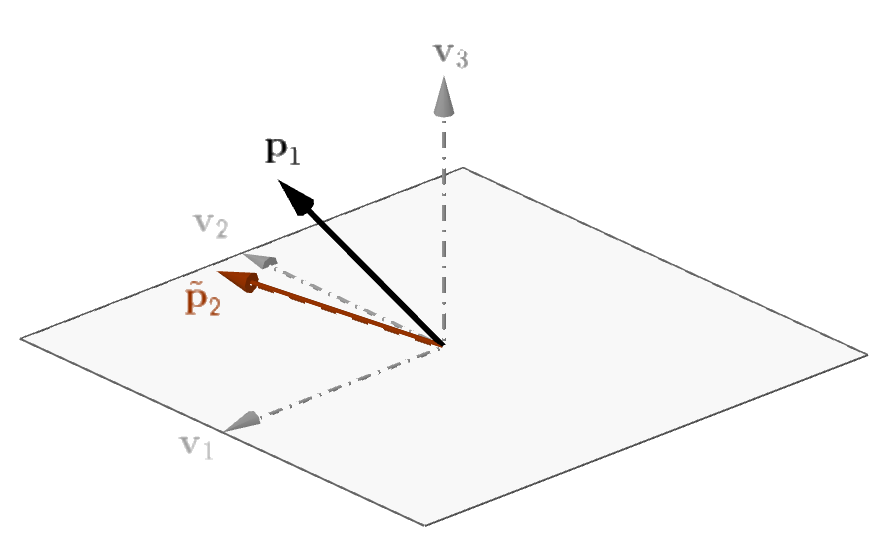}
        \caption{}
        \label{fig: replay nullspace close to v2}
    \end{subfigure}\hfill
    \begin{subfigure}{0.25\textwidth}
        \includegraphics[width=\linewidth]{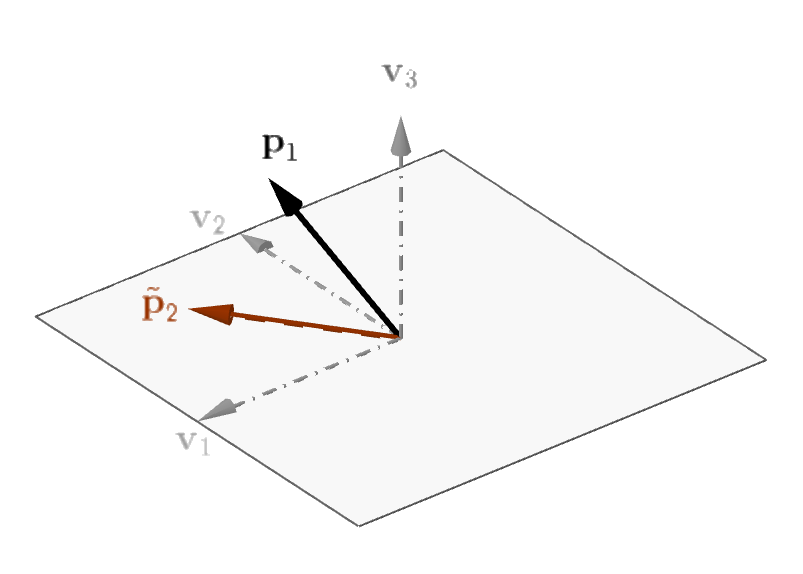}
        \caption{}
        \label{fig: replay nullspace midway from v2}
    \end{subfigure}\hfill
    \begin{subfigure}{0.25\textwidth}
         \includegraphics[width=\linewidth]{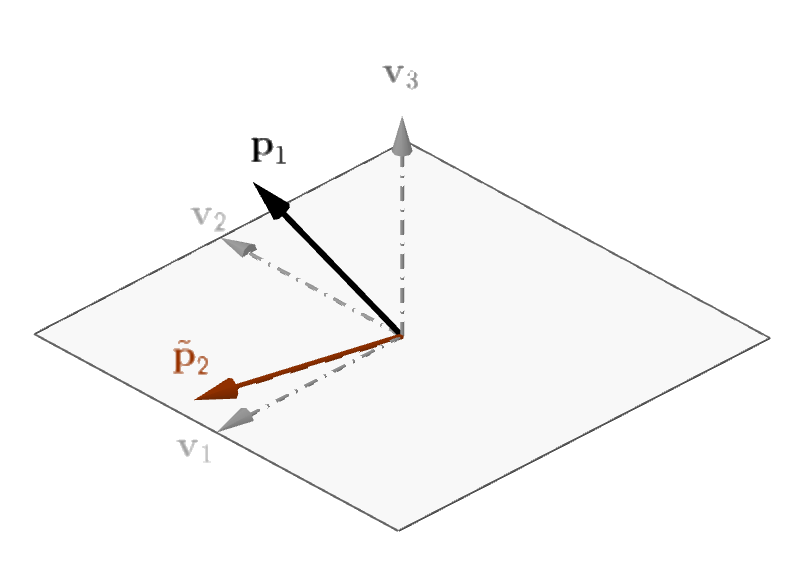}
        \caption{}
        \label{fig: replay nullspace away from v2}
    \end{subfigure}\hfill
   
    \caption{ An illustration of forgetting in the average case construction. The red plane in (a) shows the null space of task $2$.
    The null space of the first task, spanned by $\bfp_1$, is in the span of $\braces{\bfv_2,\bfv_3}$ and very close to $\bfv_2$. Without replay, the angle between $\bfv_2$ and $\bfp_1$ determines forgetting , while with replay of one sample, the angle between $\Tilde{\bfp}_2$ and $\bfp_1$ determines forgetting. In displays \ref{fig: replay nullspace close to v2} and \ref{fig: replay nullspace midway from v2} where $\Tilde{\bfp}_2$ is not too far from $\bfv_2$, forgetting would increase with replay, while in \ref{fig: replay nullspace away from v2}, it would decrease.
    }
    \label{fig: replay task 2 null space}
\end{figure}

For this intuition, we focus on the lower dimensional case ($d=3$).
Fix an orthonormal basis $\bfv_1, \bfv_2, \bfv_3$ of $\reals^3$ and consider two tasks in $\reals^3$, where the first and second tasks' null spaces has rank one and two respectively. 
Let the unit vector $\bfp_1$ span the first task's null space. $\bfp_1$  is chosen such that it is in the span of $\braces{\bfv_2, \bfv_3}$, and is very close to $\bfv_2$. The second task's null space is spanned by $\braces{\bfv_1, \bfv_2}$.
See \cref{fig: avg case null spaces} for an illustration of the null spaces.
Replaying a sample in this setting would reduce the rank of the second task's null space. 
But this doesn't necessarily mean that the forgetting will be smaller. It is known that forgetting depends on the angle between the task null spaces in a non-monotonic way; see \cref{sec: explanation for angle between null spaces} for more details. Initially, this angle is the angle between $\bfp_1$ and $\bfv_2$, which is very small.
After replay, the second task's null space will be reduced to a one dimensional null space, spanned by $\Tilde{\bfp}_2$.
Displays (b) and (c) in \cref{fig: replay task 2 null space} show some possible scenarios for $\Tilde{\bfp}_2$ where forgetting would increase, since the angle between $\bfp_1$ and $\Tilde{\bfp}_2$ would be slightly larger than the angle between $\bfp_1$ and $\bfv_2$, but not much larger. 
\cref{fig: replay nullspace away from v2} on the other hand shows a scenario where forgetting would decrease with replay since the angle between $\Tilde{\bfp}_2$ and $\bfp_1$ is much larger than the angle between $\bfp_1$ and $\bfv_1$.
The construction in \cref{thm: avg case counter example} is such that replay of most samples would result in cases like (b) and (c) in \cref{fig: replay task 2 null space}.

\section{Proofs}

\subsection{Proof of worst case results}\label{sec: worst case proofs}

\begin{proofof}{\cref{thm: worst case counter example}} \label{proof: worst case theorem}
    We construct a sequence of tasks where all the examples have unit norm. 
  Fix an arbitrary orthonormal basis $\bfv_1, \dots, \bfv_d$ for $\reals^d$ and consider the subspaces spanned by $\braces{\bfv_1, \bfv_2, \bfv_3}$ and $\sW = \lspan \braces{\bfv_4, \dots , \bfv_d}$. 
  Let
  \begin{align}
    \vx_1 = & \bfv_1,\\
    \vx_2 = & \frac{1}{2 \sqrt{2}} \bfv_1 + \frac{1}{2 \sqrt{2}} \bfv_2 + \frac{\sqrt{3}}{2} \bfv_3,\\
    \vx_3 = & \bfv_3,
\end{align}

For each task $ 1 \le t \le T-1$, let $\mX'_t$ be a matrix containing a set of samples that are in the span of $\sW$. That is, span of each $\mX'_t$ is a subset of $\sW$. The number of samples in each $\mX'_t$ does not affect our construction.

For the first $T-2$ tasks, set 

\begin{align}
    \mX_t = \begin{bmatrix}
        \vx_1^\top \\
        \mX'_t
    \end{bmatrix}.
\end{align}
For the last two tasks we have 
\begin{align}
    \mX_{T-1} = \begin{bmatrix}
        \vx_1^\top\\
        \vx_2^\top\\
        \mX'_{T-1}
    \end{bmatrix},
\end{align}

and finally for the last task
\begin{align}
    \mX_T = \begin{bmatrix}
        \vx_3^\top \\
        {\vw'_1}^\top \\
        \vdots \\
        {\vw'_{d-3}}^\top
    \end{bmatrix},
\end{align}
where $\vw'_1, \dots \vw'_{d-3}$ span $\sW$. 
Let $\bfu = \sqrt{\frac{6}{7}} \bfv_2 - \frac{1}{\sqrt{7}} \bfv_3 $ be a vector that is orthogonal to $\vx_1$ and $\vx_2$.
We pick $\vw^*$ such that $ a \defeq \bfu^\top \vw^*$ is bounded away from zero.
To compute forgetting without replay, we first compute 
$ \mP_{T} \mP_{T-1} \dots \mP1 \vw^*$. For every $t < T-1$, we decompose the null space $\mP_t = \mP'_{t} + \Pbar$, where $\Pbar$ is a projection into the span of $\bfv_2, \bfv_3$ (which are orthogonal to $\vx_1$) and $\mP'_t$ gives a projection of the null space $\mP_t$ into $\sW$. Note that $\mP'_t$ and $\Pbar$ are projections into orthogonal subspaces, hence $\mP'_t \Pbar  = \Pbar \mP'_t = 0$.
For $t < T-2$, we can write $\mP_{t+1} \mP_t = (\mP'_{t+1} + \Pbar) (\mP'_t + \Pbar)= \mP'_{t+1} \mP'_t + \Pbar$. Then we have 
\begin{align}
    \mP_{T-2} \dots \mP_2 \mP_1 \vw^* = \mP'_{T-2} \dots \mP'_2 \mP'_1 \vw^* +  \Pbar \vw^* 
\end{align}
For the second to last task, we have $\mP_{T-1} = \bfu \bfu^\top + \mP'_{T-1}$, so $ \mP_{T-1} \dots \mP_1 \vw^* = \bfu \bfu^\top \Pbar \vw^*  + \mP'_{T-1} \dots \mP'_1 \vw^*$. Since $\vw'_1, \dots \vw'_{d-3}$ span $\sW$, the last task's null space $\mP_T$ is a projection into the subspace spanned by $\bfv_1, \bfv_2$, therefore, $ \mP_T  \mP'_{T-2} \dots \mP'_2 \mP'_1 \vw^* = 0$ and consequently 
\begin{align}
    \mP_{T}  \mP_{T-1} \dots \mP_1 \vw^* = \mP_T \bfu \bfu^\top \Pbar \vw^*.
\end{align}
Since $\bfu$ is fully in the span of $\Pbar$, $\bfu \bfu^\top \Pbar = \bfu \bfu^\top$. 
Finally, the forgetting is
\begin{align}
    \frac{1}{T-1}\sum_{t=1}^{T - 1} \twonorm{\mX_t \mP_T \bfu \bfu^\top  \vw^* }^2.
\end{align}

We first compute forgetting on the samples $\vx_1, \vx_2$ and $\vx_3$. 
Since $\vx_3$ is included in the last task, $\vx_3^\top \mP_T \bfu \bfu^\top \vw^* = 0$, and since $\vx_1^\top \mP_T = \vx_1^\top$, we have $ \vx_1^\top \mP_T \bfu \bfu^\top \vw^* = \vx_1^\top \bfu \bfu^\top \vw^* = 0$. Now, for $\vx_2$ we have 
\begin{align} \label{eq: worst case x2}
    \vx_2^\top  \mP_T \bfu \bfu^\top \vw^* = & \begin{bmatrix}
        \frac{1}{2 \sqrt{2}} & \frac{1}{2 \sqrt{2}}
    \end{bmatrix} \begin{bmatrix}
        \bfv_1^\top \\
       \bfv_2^\top
    \end{bmatrix} \bfu \bfu^\top \vw^* 
    \\ = & \frac{1}{2 \sqrt{2}}\bfv_2^\top \bfu \bfu^\top  \vw^* 
    = 
    \frac{\sqrt{3}}{ 2\sqrt{7}} \cdot a.
\end{align}
The last equality follows from $ \bfv_2^\top \bfu = \sqrt{\frac{6}{7}}$ and definition of $a$.
The rest of the samples are in $\sW$, and since the samples in $\mX_T$ span $\sW$, we have that for all $i$, $\mX'_t \mP_T  = 0$, that is, they make no  contribution to forgetting. So the only sample that contributes to forgetting is $\vx_2$, which only occurs in task $T-1$. Then plugging in \cref{eq: worst case x2} we have 
\begin{align}
    \frac{1}{T-1}\sum_{t=1}^{T - 1} \twonorm{\mX_t \mP_T \bfu \bfu^\top  \vw^* }^2 
     = &
    \frac{1}{T-1} \paren{\vx_2^\top \mP_T \bfu \bfu^\top  \vw^* }^2
    \\ = &
    \frac{3}{28 (T-1)} \cdot a^2, 
\end{align}
which is of order $1/T$ as long as $a$ is bounded away from zero. 

Now, we compare this to forgetting with replay of the sample $\vx_2, y_2$, which is the only sample that contributed to forgetting. In the replay scenario, $\vx_2$ is combined with $\mX_t$ to get $\Xtil_T$, consequently the null space $ \Ptil_T = \frac{1}{2} (\bfv_1 - \bfv_2)(\bfv_1 - \bfv_2)^\top$. Recall that without replay $ \mP_T = \begin{bmatrix}
    \bfv_1, \bfv_2
\end{bmatrix} \begin{bmatrix}
    \bfv_1^\top \\
    \bfv_2^\top.
\end{bmatrix}$

Now we compute the forgetting $ \frac{1}{T-1}\sum_{t=1}^{T - 1} \twonorm{\mX_t \Ptil_T \bfu \bfu^\top  \vw^* }^2$. 
Similar to the no replay case, we can see that for all $i$, $\mX'_t \Ptil_T = 0$ and $\vx_3 \Ptil_T = 0$, so these samples don't contribute to forgetting. Additionally, since $\vx_2$ has just been replayed, it is in the span of $\Xtil_T$ and doesn't contribute to forgetting. We are left with 
\begin{align}
    \vx_1^\top  \Ptil_T \bfu \bfu^\top \vw^* = & \frac{1}{2}  (\bfv_1 - \bfv_2)^\top \bfu \bfu^\top \vw^* 
    =  - \frac{6}{28} a
\end{align}
Forgetting with replay is then $\frac{T-1}{T-1} \paren{\vx_1^\top  \Ptil_T \bfu \bfu^\top \vw^* }^2 = \frac{9}{ 196 } a^2$, since $\vx_1$ appeared in all the first $T-1$ tasks. 
Comparing $\frac{3 a^2}{28 \; (T-1)}$ to $\frac{9 a^2}{ 196 } $, we can see that as $T \rightarrow \infty$, forgetting vanishes without replay, while with replay, it is  a constant.
\end{proofof}

\subsection{Lower dimensional average case results}

\begin{theorem}[Average case replay]\label{thm: avg case counter example}
Suppose that assumptions \ref{assump: overparameterized lr}, \ref{assump: realizability}, and \ref{assump: avg case subspace covered} hold.
For every $\vw^* \in \reals^3$ where $\twonorm{\vw^*} \le 1$, there exists a sequence of two task subspaces, such that replaying a randomly chosen sample from the first task's samples increases expected forgetting.
That is 
\begin{align}
    \E \brackets{F_{S'}(\vw_2)} < \E \brackets{F_{S'}(\Tilde{\vw}_2)},
\end{align}
where $\Tilde{\vw}_2$ is the iterate after the second task with replay.
\end{theorem}

\begin{proofof}{\cref{thm: avg case counter example}} \label{proof: thm avg case}
We use the following claim  which simplifies the form of expected forgetting in the two task case with replay.  Proof of this claim is given in \cref{sec: claim proofs}

\begin{claim}\label{claim: average forgetting}
    Suppose that we have a sequence $S$ of two tasks in the average case setting (as described in \cref{sec: average case results}). Let $R$ be a set of $m \le n_1$ randomly chosen (without replacement) indices of the samples to be replayed from the first task.  Let $\mP_1, \mP_2$ be the task null spaces and $\Ptil_2 = \Ptil_2(\braces{x_{1, j}}_{j \in R})$ be the null space of the second task with replay. Then the expected forgetting with respect to test samples and with replay of $m$ samples from the first task can be simplified to
    \begin{align}
    \E \brackets{ F_{S'}(\Tilde{\vw}_2)} =
       \; \E \brackets{\twonorm{\mPi_1 \Ptil_2 \mP_1 \vw^*}^2},
    \end{align}
    where the expectation is over the randomness of $\Ptil_2$. 

\end{claim}

Given $\vw^*$, we can pick orthonormal basis $W_1$ and $W_2$ for the subspaces, $n_1$, and $n_2$.
Also recall the sample generation process described in \cref{eq: avg case sample generation}.  
We start with describing the two task subspaces. 
We first fix $\epsilon = \sqrt{\frac{1}{63}}$, though other small values of $\epsilon$ would work as well.
Then we fix an orthonormal basis $\bfv_1, \bfv_2, \bfv_3$ such that $\twonorm{\mP_1 \vw^*} > 0$, where $\mP_1$ is defined in \cref{eq: avg case P1 def}. 
The first task is spanned by orthonormal vectors $\bfv_1, \bfu$, where $ \bfu = \epsilon \bfv_2 + \sqrt{1 - \epsilon^2} \bfv_3$ for some $ 1> \epsilon > 0$ that will be fixed later. That is 
\begin{align}
    \mPi_1 = \mW_1 \mW_1^\top =  \begin{bmatrix}
        \bfv_1, \bfu
    \end{bmatrix}
    \begin{bmatrix}
        \bfv_1^\top \\
        \bfu^\top 
    \end{bmatrix}.
\end{align}
This leads to the following one dimensional null space for the first task 
\begin{align} \label{eq: avg case P1 def}
    \mP_1 = (\sqrt{1 - \epsilon^2} \bfv_2 - \epsilon \bfv_3) (\sqrt{1 - \epsilon^2} \bfv_2 - \epsilon \bfv_3)^\top.
\end{align}

The second task is spanned by $\bfv_3$, leading to the null space 
\begin{align}
    \mP_2 = 
    \begin{bmatrix}
        \bfv_1, \bfv_2
    \end{bmatrix}
    \begin{bmatrix}
        \bfv_1^\top \\
        \bfv_2^\top 
    \end{bmatrix}.
\end{align}

Now we compute forgetting without replay using \cref{eq: expected forgetting}. So we start with computing
\begin{align} \label{eq: forget error no replay}
    \mPi_1 \mP_2 \mP_1 \vw^*  = & \bfv_1 \bfv_1^\top \mP_2 \mP_1 \vw^* + 
    \bfu \bfu^\top \mP_2 \mP_1 \vw^* 
    \\
    = & \bfu \bfu^\top \mP_2 \mP_1 \vw^*.
\end{align}
The last equality holds since $ \bfv_1^\top \mP_2  = \bfv_1^\top$ and $\bfv_1^\top \mP_1 = 0$.
Let $a \defeq ((\sqrt{1 - \epsilon^2}) \cdot \bfv_2 - \epsilon \cdot \bfv_3)^\top \vw^*  $.
We have 
\begin{align}
    \bfu \bfu^\top \mP_2 \mP_1 \vw^* = & \bfu \begin{bmatrix}
        0, \epsilon
    \end{bmatrix}
    \begin{bmatrix}
        \bfv_1^\top \\
        \bfv_2^\top
    \end{bmatrix}
    \mP_1 \vw^* 
    \\ = & 
    \epsilon \cdot a \cdot  \bfu  \bfv_2^\top (\sqrt{1 - \epsilon^2} \bfv_2 - \epsilon \bfv_3) 
    \\ = & \epsilon \cdot \sqrt{1 - \epsilon^2} \cdot a \cdot \bfu.
\end{align}
Forgetting without replay is 
\begin{align}
  \twonorm{\mPi_1 \mP_2 \mP_1 \vw^*}^2 
    = &  \cdot \epsilon^2 \cdot ( 1- \epsilon^2) \cdot a^2 \cdot  \twonorm{\bfu}^2 
    \\ = &  \label{eq: forgetting without replay}
    \cdot \epsilon^2 \cdot ( 1- \epsilon^2) \cdot a^2.
\end{align}
Note that forgetting on the last task is always zero.
To compute forgetting with replay of one sample, we need to understand the distribution of $\Tilde{\mP}_2$ first. 
Let $\rmX_{1\ranj}$ be a randomly chosen sample from the first task. By  \cref{eq: avg case sample generation},
$ \rmX_{1\ranj} = \mW_1 \rmZ_{1\ranj}$ where $\rmZ_{1j} \sim \normal(0, \frac{\id_2}{2})$ and $\ranj \in_R [n_t]$ is an index that is picked uniformly at random from the set $[n_t]$. It's important here to note that 
$\ranj$ and $\rmZ_{1\ranj}$ are independent and $\rmZ_{1i}$ are iid. 
Then we have 
\begin{align} \label{eq: distribution of the randomly chosen sample }
    p (\rmZ_{1\ranj}) = & \sum_{j=1}^{n_t} p(\rmZ_{1\ranj} \mid \ranj = j) p(\ranj = j)
    \\ = & 
    \frac{1}{n_t} \sum_{j=1}^{n_t} p(\rmZ_{1\ranj} \mid \ranj = j)
    \\ = & 
    \frac{1}{n_t} \sum_{j=1}^{n_t} p(\rmZ_{1j} )
    = p (\rmZ_{11}),
\end{align}
and consequently $\rmZ_{1\ranj} \sim \normal(0, \frac{\id_2}{2})$ and we can write 
\begin{align}
    \rmX_{1\ranj} = \mW_1 \rmZ_{1\ranj} = \frac{1}{2} \paren{\alpha_1 \bfv_1 + \alpha_2 \bfu},
\end{align}
where $\frac{1}{2}\alpha_1, \frac{1}{2}\alpha_2 \sim \normal(0, \frac{1}{2})$ are the two iid coordinates of $\rmZ_{1\ranj}$.

$\Ptil_2$ is a projection onto the null space of the space spanned by task two samples, which have the form $\rmX_{2j}= \bfv_3 \rmZ_{2j}$, plus $\rmX_{1\ranj}$. 
In another words, since all the samples for task $2$ are colinear with $\bfv_3$, $\Ptil_2$ is a projection onto the null space of linear span of $\braces{\bfv_3, \alpha_1 \bfv_1 + \alpha_2 \bfu}$.
Since $\alpha_1 \bfv_1 + \alpha_2 \bfu =  \alpha_1 \bfv_1 + \alpha_2 \epsilon \bfv_2 + \alpha_2 \sqrt{1 - \epsilon^2} \bfv_3$, we have that 
\begin{align}
     & \lspan \braces {\bfv_3, \alpha_1 \bfv_1 + \alpha_2 \epsilon \bfv_2 + \alpha_2 \sqrt{1 - \epsilon^2} \bfv_3 }
    \\ & = 
    \lspan \braces{\bfv_3, \alpha_1 \bfv_1 + \alpha_2 \epsilon \bfv_2 }
\end{align}

Then $\Ptil_2$ is a projection onto a one dimensional vector space that is orthogonal to $\bfv_3$ and $\alpha_1 \bfv_1 + \alpha_2 \epsilon \bfv_2$.  We can write
\begin{align}
    \Ptil_2 =  \frac{1}{z} \cdot \paren{
        \alpha_2 \epsilon \bfv_1 - \alpha_1 \bfv_2 }
         \paren{
        \alpha_2 \epsilon \bfv_1 - \alpha_1 \bfv_2 }^\top ,
\end{align}
where $z \defeq \epsilon^2 \alpha_2^2 + \alpha_1^2$ is a  normalizing constant.
Next we compute each of the terms in 
\begin{align}
    \mPi_1 \Ptil_2 \mP_1 \vw^* = & \bfv_1 \bfv_1^\top \Ptil_2 \mP_1 \vw^* + 
    \bfu \bfu^\top \Ptil_2 \mP_1 \vw^* .
\end{align}

We have 
\begin{align}
     \bfv_1 \bfv_1^\top \Ptil_2 \mP_1 \vw^* 
     = & 
     \frac{1}{z} \cdot \bfv_1 \bfv_1^\top 
     \paren{
        \alpha_2 \epsilon \bfv_1 - \alpha_1 \bfv_2 }
         \paren{
        \alpha_2 \epsilon \bfv_1 - \alpha_1 \bfv_2 }^\top 
        \mP_1 \vw^*
     \\ = &
     \frac{\alpha_2 \; \epsilon}{z} \cdot \bfv_1  
     \paren{\alpha_2 \epsilon \bfv_1 - \alpha_1 \bfv_2 }^\top
     \mP_1 \vw^*
     \\ = &
    \frac{\alpha_2 \; \epsilon}{z} \cdot \bfv_1  
     \paren{\alpha_2 \epsilon \bfv_1 - \alpha_1 \bfv_2 }^\top
     (\sqrt{1 - \epsilon^2} \bfv_2 - \epsilon \bfv_3) 
     \\ & (\sqrt{1 - \epsilon^2} \bfv_2 - \epsilon \bfv_3)^\top
     \vw^*
     \\ = & 
     - \frac{\alpha_1 \; \alpha_2 \; \epsilon \; \sqrt{1 - \epsilon^2}}{z} \cdot a \cdot \bfv_1,
\end{align}

and 

\begin{align}
 \bfu \bfu^\top \Ptil_2 \mP_1 \vw^* = &  - \frac{\alpha_1 \epsilon}{z} \cdot \bfu \; 
 \paren{\alpha_2 \epsilon \bfv_1 - \alpha_1 \bfv_2}^\top \mP_1 \vw^* 
 \\ = & 
    \frac{\alpha_1^2 \; \epsilon \; \sqrt{1 - \epsilon^2}}{z} \cdot a \cdot \bfu .
\end{align}

Since $\bfu$ and $\bfv_1$ are orthogonal to each other, we can write 
\begin{align}
    \twonorm{\mPi_1 \Ptil_2 \mP_1 \vw^*}^2 = &
    \twonorm{\bfu \bfu^\top \Ptil_2 \mP_1 \vw^*}^2
    + 
    \twonorm{ \bfv_1 \bfv_1^\top \Ptil_2 \mP_1 \vw^*}^2
    \\ = & 
     \alpha_1^4 \epsilon^2 ( 1  - \epsilon^2) \frac{a^2}{z^2}
     +
    \alpha_1^2 \alpha_2^2 \; \epsilon^2 (1 - \epsilon^2)
     \; \frac{a^2}{z^2} 
    \\ = & 
    \paren{\frac{\alpha_1^2 \alpha_2^2 + \alpha_1^4}{z^2}}
    \epsilon^2 ( 1  - \epsilon^2) a^2 .
\end{align}
By \cref{claim: average forgetting}, the expected forgetting with replay is 
\begin{align}
     \E \brackets{\twonorm{\mPi_1 \Ptil_2 \mP_1 \vw^*}^2}
     =  
     \E \brackets{\frac{\alpha_1^2 \alpha_2^2 + \alpha_1^4}{z^2}} \epsilon^2 ( 1  - \epsilon^2) a^2.
\end{align}
We compare this to expected forgetting without replay, which is $\epsilon^2 ( 1  - \epsilon^2) a^2$, and show that there exists $\epsilon$ such that 
\begin{align} \label{eq: lower bound on expectation}
     \E \brackets{\frac{\alpha_1^2 \alpha_2^2 + \alpha_1^4}{z^2}} > 1.
\end{align}

Note that by definition, $a^2 = \twonorm{\mP_1 \vw^*}^2$ and the orthonormal basis $\bfv_1, \bfv_2, \bfv_3$ can be chosen such that  
$a^2 > 0$.
.

By definition of $z$, can write $\frac{\alpha_2^2}{z} = (1 - \frac{\alpha_1^2}{z}) \cdot \frac{1}{\epsilon^2}$ and simplify the first term inside expectation to 
$
    \frac{\alpha_1^2 \alpha_2^2}{z^2} = 
    \frac{\alpha_1^2}{z} \cdot \paren{1 - \frac{\alpha_1^2}{z}} \cdot \frac{1}{\epsilon^2}.
$

Let $\alpha'^2_1 = \frac{\alpha_1^2}{z}$, then rewriting 
\cref{eq: lower bound on expectation}, we have picked $\epsilon$ such that by \cref{claim: lower bound on expectation for two task average case counter example}
\begin{align}
    \E \brackets{ \alpha'^2_1 \cdot ( 1 - \alpha'^2_1) \frac{1}{\epsilon^2} + \alpha'^4_1 } > 1.
\end{align}

  We can then conclude that replay has increased average forgetting. 

\begin{claim}\label{claim: lower bound on expectation for two task average case counter example}
    Let $ \alpha_1, \alpha_2 \sim \normal(0, 1)$, and ${\alpha'_1}^2 
    = \frac{\alpha_1^2}{\frac{\alpha_2^2}{63} + \alpha_1^2}$.  Then 
    \begin{align}
        \E \brackets {63 {\alpha'_1}^2 - 62 {\alpha'_1}^4} \ge 1.4 .
    \end{align}
\end{claim}
 Proof of this claim is given in \cref{proof: proof of rough lower bound on alphas}.

\end{proofof}

\subsection{Proof of \cref{thm: average case high dimensional construction}}

\begin{proofof}{\cref{thm: average case high dimensional construction}} \label{proof: average case high dimensional}
The construction for this higher dimensional case is an extension of the lower dimensional one. We start by describing the construction, and then calculate forgetting with and without replay. Calculating expected forgetting with replay is more involved. 
\paragraph{Construction}
Let $d$ be the input dimension.
The first task is a $d-1$ dimensional subspace spanned by $\braces{\bfu, \bfv_1, \bfv_3, \dots, \bfv_{d-1}}$, where $\bfu = \epsilon \bfv_2 + \sqrt{1- \epsilon^2} \bfv_d$, for some $0 < \epsilon < \frac{1}{2}$. So columns of $\mW_1$ consist of $\bfu, \bfv_1, \bfv_3, \dots, \bfv_{d-1}$.  The subspace for the second task is spanned by $\bfv_d$.

\paragraph{Simplifying the forgetting expression}
We can check that $\vu_\perp =\sqrt{1 - \epsilon^2} \cdot \bfv_2 - \epsilon \cdot  \bfv_d$ spans the null space of the first task, that is $\mP_1 = \vu_\perp \vu_\perp^\top$. 
Plugging this in the forgetting expression (\cref{eq: expected forgetting}) we get
\begin{align} \label{eq: general forgetting simplification construction}
    \twonorm{\mPi_1 \mP_2 \mP_1 \vw^*}^2 & = \twonorm{(\id - \mP_1) \mP_2 \mP_1 \vw^*}^2  = \twonorm{\mP_2 \mP_1 \vw^*}^2  - \twonorm{\mP_1 \mP_2 \mP_1 \vw^*}^2 
    \\ & = a^2 \twonorm{\mP_2 \vu_\perp}^2 - a^2 \twonorm{ \vu_\perp \vu_\perp^\top \mP_2 \vu_\perp}^2 
    \\ \label{eq: general forgetting form higher dimensional construction} & = 
    a^2 \paren{ \twonorm{\mP_2 \vu_\perp}^2 - \twonorm{\mP_2 \vu_\perp}^4 },
\end{align}

where $ a = \vu_\perp^\top \vw^*$, and last equality follows from  $ \twonorm{ \vu_\perp \vu_\perp^\top \mP_2 \vu_\perp}^2  = (\vu_\perp^\top \mP_2 \vu_\perp)^2 = \twonorm{\mP_2 \vu_\perp}^4$. 
It is then easy to see that forgetting is maximized when $\twonorm{\mP_2 \vu_\perp}^2 = \frac{1}{2}$, as shown in \cite{evron22a}.

\paragraph{Forgetting without replay}
To find forgetting without replay, it would suffice to compute \cref{eq: general forgetting form higher dimensional construction}.
The null space of the second task is spanned by $\bfv_1, \dots, \bfv_{d-1}$, so projection of $\vu_\perp =\sqrt{1 - \epsilon^2} \cdot \bfv_2 - \epsilon \cdot  \bfv_d$ into this null space would just be $ \mP_2 \vu_\perp = \sqrt{1 - \epsilon^2} \cdot \bfv_2 $.
Then forgetting without replay would be 
\begin{align}
    \twonorm{\mPi_1 \mP_2 \mP_1 \vw^*}^2  = a^2 \cdot \paren{ (1 - \epsilon^2) - (1- \epsilon^2)^2 } = a^2  \cdot \epsilon^2 \cdot (1 - \epsilon^2).
\end{align}

\paragraph{Forgetting with replay}
By \cref{claim: average forgetting}, to find expected forgetting with replay, we need to find $\E \brackets {\twonorm{\mPi_1  \Ptil_2 \mP_1 \vw^*}^2}$, where the expectation is over the randomness in $\Ptil_2$. 
We can use the same steps as in \cref{eq: general forgetting simplification construction} to get 
\begin{align} \label{eq: replay avg case hd}
    \twonorm{\mPi_1  \Ptil_2 \mP_1 \vw^*}^2 = a^2 \cdot \twonorm{\Ptil_2 \vu_\perp}^2 \cdot \paren{ 1 - \twonorm{\Ptil_2 \vu_\perp}^2}.
\end{align}

To understand $ \twonorm{\Ptil_2 \bfu_\perp}^2$, we first focus on the span of the second tasks samples combined with replay samples. Note that the $m$ replay samples from the first task have the form $ \alpha_1 \bfv_1 + \alpha_2 \vu+ \dots + \alpha_{d-1} \bfv_{d-1}$, where $\alpha_i \sim \normal(0, \frac{1}{d-1})$. Since the second task is only spanned by $\bfv_d$, we can make the replay samples orthogonal to $\mPi_2$ by subtracting this component from each sample. Then span of the second task, including the  $m$ replay samples is 
\begin{align}
    \lspan \braces{\bfv_d, \Ztil_{11}, \dots, \Ztil_{1m}}
\end{align}

where $\Ztil_{1i} = \alpha_1^{(i)} \cdot \bfv_1 + \alpha_2^{(i)} \cdot  \epsilon \bfv_2   + \dots + \alpha_{d-1}^{(i)} \bfv_{d-1}$, and $\alpha_j^{(i)} \sim \normal(0, 1)$.
That is, $\bfu$ is replaced by $\epsilon \bfv_2$.
We have ignored the multiplicative factor of $\frac{1}{d-1}$, since that would not affect the span. Null space of $\lspan \braces{\bfv_d, \Ztil_{11}, \dots, \Ztil_{1m}}$ is orthogonal to $\bfv_d$, so $ \Ptil_2 \bfu_\perp = \Ptil_2  \paren{\sqrt{1 - \epsilon^2} \cdot \bfv_2 - \epsilon \cdot  \bfv_d} = \sqrt{1- \epsilon^2 } \cdot \Ptil_2 \bfv_2$. 
Let $\Pitil_2$ be the orthogonal projection into the span of $\braces{ \Ztil_{11}, \dots, \Ztil_{1m}}$, then we can write $ \Ptil_2 = \id - (\bfv_d \bfv_d^\top + \Pitil_2)$. Next, we focus on $\Pitil_2 \bfv_2$.
Without loss of generality, we can assume that $\bfv_1, \dots , \bfv_{d-1}$ are canonical basis vectors, and $\Ztil_{1i} \sim \normal(0, \Sigma)$ where $ \Sigma$ is a $d-1$ dimensional diagonal matrix,
\begin{align}
    \Sigma = \begin{bmatrix}
        1 &  & \\
         & \epsilon^2  & \\
         &  & \id_{d-3}
    \end{bmatrix}.
\end{align}
Equivalently, we could write $\Ztil_{1i} = \Sigma^{1/2} \Zpr_{1i}$, where $\Zpr_{1i} \sim \normal(0, \id_{d-1})$. Collecting what we have so far we get 
\begin{align}
    \Ptil_2 \bfu_\perp =  \sqrt{1- \epsilon^2 } \paren {\id - \bfv_d \bfv_d^\top - \Pitil_2} \bfv_2 =  \sqrt{1- \epsilon^2 } \paren {\id - \Pitil_2} \bfv_2,
\end{align}
since $\bfv_d$ is orthogonal to $\bfv_2$.
We can write 
\begin{align} \label{eq: projection to random samples}
    \twonorm{\Ptil_2 \bfu_\perp}^2 = (1- \epsilon^2) \paren{\twonorm{\bfv_2}^2 - \twonorm{\Pitil_2 \bfv_2}^2} = 
    (1- \epsilon^2) \paren{ 1 - \twonorm{\Pitil_2 \bfv_2}^2}.
\end{align}

To bound expectation of the term given in \cref{eq: replay avg case hd}, we first give high probability upper and lower bounds for $\twonorm{\Pitil_2 \bfu_\perp}^2$.

Let $\Ztil$ be the $m \times d-1$ dimensional matrix that has $\Ztil_{1i}$ as rows, and similarly, let $\Zpr$ be the $m \times d-1$ dimensional matrix with $\Zpr_{1i}$ in the rows. Then we can write $ \Ztil = \Zpr \Sigma^{1/2}$.

One way to write the projection is  $\Pitil_2 = \Ztil^\top \paren{\Ztil \Ztil^\top}^{-1} \Ztil$, this can be checked by writing the singular value decomposition of $\Ztil$. Going back to calculating 
$\twonorm{\Pitil_2 \bfv_2}^2$,

\begin{align}
    \twonorm{\Pitil_2 \bfv_2}^2 = 
    \bfv_2^\top \Ztil^\top \paren{\Ztil \Ztil^\top}^{-1} \Ztil \bfv_2 & = 
    \bfv_2^\top \Sigma^{1/2} \Zpr^\top \paren{\Zpr \Sigma \Zpr^\top }^{-1} \Zpr \Sigma^{1/2} \bfv_2.
\end{align}

Note that $ \Sigma^{1/2} \bfv_2 = \epsilon \bfv_2$,  and  since $ \Sigma \succeq \epsilon^2 \id_{d-1}$, $ \paren{\Zpr \Sigma \Zpr^\top }^{-1} \preceq \frac{1}{\epsilon^2}  \cdot \paren{\Zpr \Zpr^\top }^{-1}$. Then 
\begin{align}
    \twonorm{\Pitil_2 \bfv_2}^2 \le  \bfv_2^\top \Zpr^\top \paren{\Zpr \Zpr^\top}^{-1} \Zpr \bfv_2.
\end{align}

On the other hand, since $ \Sigma \preceq \id_{d-1}$, $    \paren{\Zpr \Zpr^\top }^{-1}  \preceq \paren{\Zpr \Sigma \Zpr^\top }^{-1} $, so

\begin{align}
     \twonorm{\Pitil_2 \bfv_2}^2 = 
     \bfv_2^\top \Sigma^{1/2} \Zpr^\top \paren{\Zpr \Sigma \Zpr^\top }^{-1} \Zpr \Sigma^{1/2} \bfv_2
     \ge \epsilon^2 \bfv_2^\top \Zpr^\top \paren{\Zpr \Zpr^\top }^{-1} \Zpr \bfv_2
\end{align}

Recall that $\Zpr$ is a $m \times (d-1)$ dimensional matrix whose entries are from a standard normal distribution, so $ \hat{\mP} \defeq \Zpr^\top \paren{\Zpr \Zpr}^{-1} \Zpr $, is a random projection into a $m$ dimensional subspace of $d-1$ dimensional space. 

Calculations above show that 
\begin{align} \label{eq: to random projection bound}
     \epsilon^2 \twonorm{\hat{\mP} \bfv_2}^2 \le \twonorm{\Pitil_2 \bfv_2}^2 \le \twonorm{\hat{\mP} \bfv_2}^2.
\end{align}

As described in \citet{Dasgupta2003AnEP}, $\hat{\mP} \bfv_2$ has the same distribution as a fixed projection (into an $m$ dimensional subspace) of a vector $\hat{\bfu}$ picked uniformly at random from the sphere. This fixed subspace can be chosen to be the first $m$ coordinates. Lemma 2.2 of \citet{Dasgupta2003AnEP}, gives concentration bounds on $\hat{\bfu}_1^2 + \hat{\bfu}_2^2 + \dots + \hat{\bfu}_m^2$. We include this lemma here using our notation for convenience.
\begin{lemma}[Lemma 2.2, \citet{Dasgupta2003AnEP}]\label{lemma: concentration of random projection}
Assume $m < d-1$. Then for $t < 1$
\begin{align}
    \Pr\brackets{ \hat{\bfu}_1^2 + \hat{\bfu}_2^2 + \dots + \hat{\bfu}_m^2 \le \frac{tm}{d-1}} \le \exp\paren{\frac{m}{2}\paren{1 - t + \ln t} },
\end{align}
and for $t > 1$
\begin{align}
    \Pr\brackets{ \hat{\bfu}_1^2 + \hat{\bfu}_2^2 + \dots + \hat{\bfu}_m^2 \ge \frac{tm}{d-1}} \le \exp\paren{\frac{m}{2}\paren{1 - t + \ln t} }.
\end{align}  
\end{lemma}

Since $\twonorm{\hat{\mP} \bfv_2}^2$ has the same distribution as $\hat{\bfu}_1^2 + \dots + \hat{\bfu}_m^2$, we can apply the lemma above to get that for $t = \frac{1}{30}$
\begin{align}
    \Pr\brackets{\twonorm{\hat{\mP} \bfv_2}^2 \le \frac{m}{30 (d-1)}} \le \exp \paren{-m},
\end{align}
and for $t=5$,
\begin{align}
    \Pr\brackets{\twonorm{\hat{\mP} \bfv_2}^2 \ge \frac{5 m}{d-1}} \le \exp \paren{- m}.
\end{align}
Using \cref{eq: to random projection bound}, we have that with probability of at least $1 -  2 \exp(-m)$
\begin{align}
   \epsilon^2  \frac{m}{30 (d-1)} 
    \le 
    \twonorm{\Pitil_2 \bfv_2}^2
    \le 
   \frac{5 m}{d-1}.
\end{align}
Recall from \cref{eq: projection to random samples} that $ \twonorm{\Ptil_2 \bfu_\perp}^2 = (1- \epsilon^2) \paren{ 1 - \twonorm{\Pitil_2 \bfv_2}^2}$, so with probability of at least $1 - 2 \exp(-m)$,

\begin{align} \label{eq: hp bound for replay projection}
     (1- \epsilon^2) \paren{ 1- \frac{5 m}{d-1} }
     \le \twonorm{\Ptil_2 \bfu_\perp}^2 
     \le (1- \epsilon^2) \paren{1 -  \epsilon^2  \frac{m}{30 (d-1)}}.
\end{align}
 Note that, we always have that $\twonorm{\Ptil_2 \bfu_\perp}^2  \le 1 - \epsilon^2$.

Define $f(x)= x (1 - x)$, from \cref{eq: replay avg case hd} we can see that expected forgetting with replay is 
\begin{align}
    \E \brackets{ \twonorm{\mPi_1  \Ptil_2 \mP_1 \vw^*}^2} = a^2 \E \brackets{ f \paren{\twonorm{\Ptil_2 \vu_\perp}^2}},
\end{align}
while expected forgetting without replay is $ a^2 f(1 - \epsilon^2)$.

Since $\epsilon < 1/2$, $1 - \epsilon^2 > 3/4$. Assume that $\frac{ m}{d-1} < \frac{1}{15}$ so that $\frac{1}{2} <  (1- \epsilon^2) \paren{ 1- \frac{5 m}{d-1} }$.
Since $ f(x)$ is a concave function that is maximized at $1/2$, it is monotonically decreasing in the interval $[1/2, 1]$. 
This means that if the bounds in \cref{eq: hp bound for replay projection} hold,
\begin{align}
    f \paren{\twonorm{\Ptil_2 \vu_\perp}^2} \ge
    f\paren{ (1- \epsilon^2) \paren{1 -  \epsilon^2  \frac{m}{30 (d-1)}} }.
\end{align}
We can lower bound 

\begin{align}
    \E \brackets{ f \paren{\twonorm{\Ptil_2 \vu_\perp}^2}} \ge 
    (1 - 2 \exp(-m))   f\paren{ (1- \epsilon^2) \paren{1 -  \epsilon^2  \frac{m}{30 (d-1)}} } 
    + 
    \exp(-m) f\paren{1 - \epsilon^2}.
\end{align}
To show that expected forgetting would increase with replay, it would suffice to show that 
\begin{align}
    (1 - 2 \exp(-m))   f\paren{ (1- \epsilon^2) \paren{1 -  \epsilon^2  \frac{m}{30 (d-1)}} } 
    + 
    \exp(-m) f\paren{1 - \epsilon^2} 
    > f(1-\epsilon^2),
\end{align}
or equivalently, 
\begin{align}
    (1 - 2 \exp(-m))   f\paren{ (1- \epsilon^2) \paren{1 -  \epsilon^2  \frac{m}{30 (d-1)}} } 
    > (1 - \exp(-m)) f(1-\epsilon^2).
\end{align}
To show this, we will argue that 
\begin{align} \label{eq: target inequality high dim avg case}
    \frac{f(1-\epsilon^2)}{f\paren{ (1- \epsilon^2) \paren{1 -  \epsilon^2  \frac{m}{30 (d-1)}} }  }
    < \frac{1 - 2 \exp(-m)}{1 -  \exp(-m)}
    = 1 - \frac{\exp(-m)}{1 -  \exp(-m)}
\end{align}

Let $\gamma \defeq \frac{m}{30(d-1)}$, then 
\begin{align}
     \frac{f(1-\epsilon^2)}{f\paren{ (1- \epsilon^2) \paren{1 -  \epsilon^2  \frac{m}{30 (d-1)}} }  } = &  \frac{f(1-\epsilon^2)}{f\paren{ (1- \epsilon^2) \paren{1 -  \epsilon^2  \gamma} }  }
     \\ = & 
     \frac{\epsilon^2 (1-\epsilon^2)}
        {(1- \epsilon^2) \paren{1 -  \epsilon^2  \gamma} \epsilon^2 (1 + \gamma - \epsilon^2\gamma)}
    \\ = & 
    \frac{1}{\paren{1 -  \epsilon^2  \gamma}(1 + \gamma - \epsilon^2\gamma)}.
\end{align}

Since $\epsilon^2 < 1/4$, we have 
\begin{align}
    \paren{1 -  \epsilon^2  \gamma}(1 + \gamma - \epsilon^2\gamma)  > 
    \paren{1 -    \frac{\gamma}{4}}(1 +  \frac{ 3 \gamma}{4}) >
    1 + \frac{\gamma}{2} - \frac{3\gamma}{16} = 1 + \frac{5 \gamma}{16}, 
\end{align}
so 
\begin{align}
    \frac{1}{\paren{1 -  \epsilon^2  \gamma}(1 + \gamma - \epsilon^2\gamma)} 
    <
    \frac{1}{ 1 + \frac{5 \gamma}{16}}
    = 1 - \frac{\frac{5 \gamma}{16}}{ 1 + \frac{5 \gamma}{16}}.
\end{align}

To show \cref{eq: target inequality high dim avg case}, it would suffice to argue that 
\begin{align}
   \frac{\frac{5 \gamma}{16}}{ 1 + \frac{5 \gamma}{16}} > 
    \frac{\exp(-m)}{1 -  \exp(-m)}
\end{align}

for some values of $m$ and $d$. Simplifying the expressions above, this would be equivalent to 
\begin{align}
    \frac{16}{5\gamma} + 1  < \exp (m) - 1.
\end{align}

Plugging in $\gamma$, we can see that this inequality would be satisfied as long as 
\begin{align}
    96 \cdot \frac{d-1}{m} + 2 < \exp(m).
\end{align}

Setting $c_1 = 120, c_2 = 15$ and $c_3 = 97$ gives the stated results.

\end{proofof}

\subsection{Proofs of claims and propositions} \label{sec: claim proofs}

\begin{proofof}{\cref{prop: expected forgetting without replay}}\label{proof: expected forgetting without replay}

Note that 
\begin{align} \label{eq: avg case cov calculation}
    \E \brackets{{\rmX'}_{tj} {\rmX'}_{tj}^\top} = & \E \brackets{\mW_t {\rmZ'}_{tj} {\rmZ'}_{tj}^\top \mW_t^\top} 
         =  
        \mW_t \frac{\id_k}{k_t} \mW_t^\top
        = \frac{1}{k_t} \; \mW_t \mW_t^\top 
        = \frac{1}{k_t} \mPi_t,
\end{align}
where $\mPi_i$ was the projection matrix into the subspace spanned by samples of task $i$.
Additionally, we can write 
\begin{align} \label{eq: task covariance}
    \E \brackets{{\rmX'}_t^\top {\rmX'}_t} = \sum_{j \in [k_t]} \E \brackets{{\rmX'}_{tj} {\rmX'}_{tj}^\top} =  \mPi_t,
\end{align}
which will be useful when we compute expected forgetting below.

Recall that $\rmZ_{tj}$ were used to generate training samples for the task (\cref{eq: avg case sample generation}).
Since any $k_t$ independent $\rmZ_{tj}$ samples are going to be linearly independent, we are guaranteed that $\rmZ_{tj}$ will have the same span as $\mW_t$ under \cref{assump: avg case subspace covered}, which states that $n_t \ge k_t$.
Consequently, the null space of each task $t$ is  $\mP_t = \id - \mPi_t$.
Then similar to \cref{eq: forgetting sequential projections}, we can write each term in the expected forgetting (with respect to test samples) as 
\begin{align} \label{eq: expected forgetting sequence of projections}
    \E \brackets {F_{S'}(\vw_T)} = & \frac{1}{T-1} \sum_{t = 1}^{T-1} \E \brackets{\twonorm{{\rmX'}_t (\vw_T - \vw^*)}^2}
     =  \frac{1}{T-1} \sum_{t = 1}^{T-1} 
    \E \brackets{\twonorm{{\rmX'}_t \mP_T \mP_{T-1} \dots \mP_1 \vw^*}^2} ,
\end{align}
where now the expectation is only over the randomness in ${X'}_t$.
Expanding the square inside the expectations in \cref{eq: expected forgetting sequence of projections} and applying \cref{eq: task covariance} we get

\begin{align}
      \E \brackets{\twonorm{\mX_t \mP_T \dots  \mP_1 \vw^*}^2}   = 
     \E \brackets{{\vw^*}^\top \mP_1 \dots \mP_T \mX_t^\top \mX_t \mP_T \dots \mP_1 \vw^*} 
      & = 
     {\vw^*}^\top \mP_1 \dots \mP_T \mPi_t \mP_T \dots \mP_1 \vw^*
    \\ & =  \twonorm{\mPi_t \mP_T \dots \mP_1 \vw^*}^2,
\end{align}
where the last equality follows from the fact that $\mPi_t$ is an orthonormal projection and $\mPi_t = \mPi_t^2$.
Plugging this back into \cref{eq: expected forgetting sequence of projections} we can write the expected forgetting as
\begin{align} 
    \E \brackets {F_{S'}(\vw_T)}
    = & \frac{1}{T-1} \sum_{t = 1}^{T-1} \twonorm{\mPi_t \mP_T \dots \mP_1 \vw^*}^2.
\end{align}
\end{proofof}

\begin{proofof}{\cref{claim: average forgetting}} \label{proof: proof of average forgetting}
Suppose that $m$ samples are randomly (without replacement) selected from the $n_1$ 
samples for task one. Alternatively, we can think of them as being fixed before seeing any samples. Let $S_1 \subseteq [n_1]$ be a randomly chosen set of indices of samples that were selected for replay. Let $\Ptil_2$ be the projection into the null space of the combined samples for the second task. 
Then $\Ptil_2 = \Ptil_2(\braces{\rmX_{1, s}}_{s \in S_1})$ is 
random, unlike $\mP_2$. The expected forgetting is

\begin{align}
    \E \brackets{F_{S'}(\Tilde{\vw}_2)} = & 
    \E \brackets{ \twonorm{{\rmX'}_1 \Ptil_2 \mP_1 \vw^*}^2}
    = \E \brackets{{\vw^*}^\top \mP_1 \Ptil_2 {\rmX'}_1^\top {\rmX'}_1 \Ptil_2 \mP_1 \vw^*}.
\end{align}
Since $\braces{\rmX_{1, s}}_{s \in S_1}$ is independent from $X'_1$ and $\E \brackets{{\rmX'}_1^\top {\rmX'}_1 } = \mPi_1$ (see \cref{eq: task covariance}), we can write the expectation above as 
\begin{align}
    \E \brackets{{\vw^*}^\top \mP_1 \Ptil_2 \mPi_1  \Ptil_2 \mP_1 \vw^*} 
    =
    \E \brackets{\twonorm{\mPi_1  \Ptil_2 \mP_1 \vw^*}^2},
\end{align}
where now the expectation is only over the randomness in $\Ptil_2$.

\end{proofof}

\begin{proofof}{\cref{claim: lower bound on expectation for two task average case counter example}} \label{proof: proof of rough lower bound on alphas}
    Without loss of generality we can assume that $\alpha_1, \alpha_2 \sim \normal (0, 1)$, as this would not change the distribution of ${\alpha'_1}^2$. Define $
    f(\alpha_1') = 63 {\alpha'_1}^2 - 62 {\alpha'_1}^4 = {\alpha'_1}^2 (63 - 62 {\alpha'_1}^2)$.
    We can lower bound the expectation by considering the following three events:
    \begin{enumerate}
        \item \label{item: unlikely gain} $ {\alpha'_1}^2 < 1/31$: under this event we use the trivial lower bound $ f(\alpha'_1) \ge 0$.
        \item \label{item: middle ground} $ 1/31 \le  {\alpha'_1}^2 \le 63/64$: then $ f(\alpha'_1) \ge 1.9 $
        \item \label{item: majority event}${\alpha'_1}^2 > 63/64$: since we always have $ {\alpha'_1}^2 \le 1$, we will use the bound $f(\alpha'_1) \ge 1$.
    \end{enumerate}
    Now we bound the probability of these events.
    Note that by symmetry, $\alpha_2^2 \le \alpha_1^2$ with probability of $1/2$, then  with would have
    \begin{align}
        {\alpha'_1}^2 
    = & \frac{\alpha_1^2}{\frac{\alpha_2^2}{63} + \alpha_1^2}
    \ge \frac{\alpha_1^2}{ \frac{\alpha_1^2}{63} + \alpha_1^2 } = \frac{63}{64}.
    \end{align}

    So the event in \cref{item: majority event} happens with probability $1/2$. Next, we argue that probability of the event in \cref{item: unlikely gain} is very small. 
    If ${\alpha'_1}^2 
    =  \frac{\alpha_1^2}{\frac{\alpha_2^2}{63} + \alpha_1^2} < 1/31$, then it must be that 
    $ 30 \cdot 63 \alpha_1^2 < \alpha_2^2$. We first argue that with high probability $ \alpha_1^2 \ge \frac{4}{30 \cdot 61}$. 
    Note that the pdf of normal distribution is upper bounded by $\frac{1}{\sqrt{2\pi}}$, so 
    \begin{align}
        \Pr \brackets{\alpha_1^2 < \frac{4}{30 \cdot 61}} 
        \le \sqrt{\frac{2 \cdot 4}{2 \pi \cdot 30 \cdot 61}} \le 0.018.
    \end{align}
    Then we have 
    \begin{align}
        \Pr \brackets{{\alpha'_1}^2 < 1/31} \le &
        \Pr \brackets{\alpha_1^2 < \frac{4}{30 \cdot 61}} + 
        \Pr \brackets{ 30 \cdot 63 \alpha_1^2 < \alpha_2^2 \mid \alpha_1^2 \ge \frac{4}{30 \cdot 61}} 
        \\ \le & 
        \; 0.018 +  \Pr \brackets{ 30 \cdot 63 \alpha_1^2 < \alpha_2^2 \mid \alpha_1^2 \ge \frac{4}{30 \cdot 61}}.
    \end{align}
    Note that 
    \begin{align}
        \Pr \brackets{ 30 \cdot 63 \alpha_1^2 < \alpha_2^2 \mid \alpha_1^2 \ge \frac{4}{30 \cdot 61}}
        \le \Pr \brackets{ 4 < \alpha_2^2 } \le 0.0001.
    \end{align}
    Now we have that $ \Pr \brackets{{\alpha'_1}^2 < 1/31} \le 0.019$. This lets us lower bound the probability of the event in \cref{item: middle ground} by $ 0.5 - 0.019 = 0.481$.
    Collecting these three bounds together we get 
    \begin{align}
        \E \brackets {63 {\alpha'_1}^2 - 62 {\alpha'_1}^4} \ge 0.481 \cdot 1.9 + 0.5 \cdot 1 \ge  1.4.
    \end{align}

\end{proofof}

\begin{proofof}{\cref{prop: far tasks no increase in forgetting}}
\label{proof: close tasks proposition}
    Note that $\lspan (\Tilde{\mP}_2) \subseteq \lspan (\mP_2)$, since combining replay samples with samples of the second task would only reduce the size of the null space. 
    We can decompose $\mP_2 = \Tilde{\mP}_2 + \Bar{\mP}_2$ where $ \Tilde{\mP}_2 \Bar{\mP}_2 = 0$. 
    Then it is easy to see that for any vector $\vw$, $\twonorm{\Tilde{\mP}_2 \mP_1 \vw}^2 \le \twonorm{\mP_2 \mP_1 \vw}^2 $.
    Define $\mA \defeq \mP_2 \mP_1$ and $\Tilde{\mA} \defeq \Tilde{\mP}_2 \mP_1$.
    This implies that 
    \begin{align}
        0 \le \vw^\top \mA^\top \mA \vw^\top - \vw^\top \Tilde{\mA}^\top \Tilde{\mA} \vw 
    \end{align}
    for every $\vw$. Since $\mA^\top \mA$ and $\Tilde{\mA}^\top \Tilde{\mA}$ are symmetric positive semidefinite matrices, for $i \in [d]$, the eigenvalues $ \eig{i}{\Tilde{\mA}^\top \Tilde{\mA}} \le \eig{i}{\mA^\top \mA}$, see for example Corollary 7.7.4 of \citet{Horn_Johnson_1985} .
    
    From \cref{eq: general forgetting simplification construction}, and using the fact that $\mP_1 \mP_2 \mP_1 = \mP_1 \mP_2 \mP_2\mP_1 $, forgetting for a fixed $\wstar$ is 
    \begin{align}
        F(\vw_2) = \twonorm{(\id - \mP_1) \mP_2 \mP_1 \wstar}^2  &= \twonorm{\mP_2 \mP_1 \wstar}^2  - \twonorm{\mP_1 \mP_2 \mP_1 \wstar}^2 
        \\ & =  \label{eq: no replay positive conditions forgetting}
        {\wstar}^\top \mA^\top \mA \wstar - \wstar^\top \paren{\mA^\top \mA }^2 \wstar.
    \end{align}
Since $\E_{\wstar}\brackets{\wstar \wstar^\top} = \id$, taking the expectation over $\wstar$ gives
\begin{align}
    \E_{\wstar} \brackets{ F(\vw_2)} & = 
    \E_{\wstar} \brackets{ \Tr\paren{\wstar^\top (\mA^\top \mA - (\mA^\top \mA)^2) \wstar} }  = \Tr \brackets{ \mA^\top \mA - (\mA^\top \mA)^2} 
    \\ &= \sum_{i=1}^d \eig{i}{\mA^\top \mA} - \eig{i}{\mA^\top \mA}^2 .
\end{align}
Similar calculations would show that with replay 
\begin{align} \label{eq: replay positive conditions forgetting}
     \E_{\wstar} \brackets{ F(\Tilde{\vw}_2)} = \sum_{i=1}^d \eig{i}{\Tilde{\mA}^\top \Tilde{\mA}} - \eig{i}{\Tilde{\mA}^\top \Tilde{\mA}}^2.
\end{align}
Since $\opnorm{\mP_2 \mP_1} \le \frac{\sqrt{2}}{2}$, all the eigenvalues of both ${\Tilde{\mA}^\top \Tilde{\mA}} $ and $\mA^\top \mA$ are less than $1/2$. 
The function $f(x) = x - x^2$ is monotonically increasing in the interval $[0,1/2]$, so we have that for each $i \in [d]$, since $ \eig{i}{\Tilde{\mA}^\top \Tilde{\mA}} \le \eig{i}{\mA^\top \mA}$
\begin{align}
    \eig{i}{\Tilde{\mA}^\top \Tilde{\mA}} - \eig{i}{\Tilde{\mA}^\top \Tilde{\mA}}^2 
    \le 
    \eig{i}{\mA^\top \mA} - \eig{i}{\mA^\top \mA}^2.
\end{align}
The equation above combined with Equations \ref{eq: replay positive conditions forgetting} and \ref{eq: no replay positive conditions forgetting} implies that $\E_{\wstar} \brackets{ F(\Tilde{\vw}_2)} \le \E_{\wstar} \brackets{ F(\vw_2)} $.

\end{proofof}

\section{Details of the Experiments}
\label{sec: experiment details}

\subsection{Continual Linear Regression Experiments}
\label{appendix: Continual LR Experiments}
The linear models were trained starting from $0$ using SGD while the neural nets were trained with Adam and randomly initialized (Glorot uniform). 
The models are trained until convergence. Unless explicitly specified otherwise, the MLPs have one hidden layer of width $128 d$ where $d$ is the input dimension.
The number of samples per task was also $10$ and $100$ for the $3$ and $50$ dimensional case respectively.

\paragraph{Training details}
Both of the linear and MLP models were trained for $7000$ epochs on each task to produce \cref{fig: replay errors 3 dim }, and $5000$ epochs in \cref{fig: replay errors 50 dim }. The batch sizes used for experiments with $3$ and $50$ dimensions are $4$ and $32$ respectively. 
The linear model in \cref{fig: replay errors 3 dim } was trained with plain SGD with learning rate $0.1$. The linear model for the higher dimensional case in \cref{fig: replay errors 50 dim } was also trained with SGD with learning rate $1$ on the first task, while for the second task, the learning rate was $0.1$ with exponential decay rate $0.8$. 

In the three dimensional case (\cref{fig: replay errors 3 dim }), the MLP was trained on the first task using Adam with learning rate $8\mathrm{e}{-4}$ and exponential decay rate $0.7$. On the second task, the learning rate was $0.001$ with exponential decay rate $0.6$.
In the $50$ dimensional case, the MLP was trained on the first task starting with learning rate $1e{-4}$ and exponential decay rate $0.6$. The starting learning rate on the second task was $0.001$ and the exponential decay rate was $0.6$.

These parameters were picked such that the training converges and training error is minimized. 
We have plotted the forgetting with higher number of independent runs, since the variance in error is quite high. Note that the statement of the average case result is on the expectation, and hence the error bars show standard mean error.  The construction of the input distribution is the same as the one given in the proof of the theorem with $\epsilon = 0.2$.

\paragraph{Replay Implementation}
Let $b$ be the batch size for training on the second task, during each training step of the second task, a random batch of $b' = \min \braces{b, m}$
many samples from the second task are combined with a random batch of $b$ samples from $(\Xmem, \Ymem)$, and they are weighted by $\frac{b}{b'}$ so that their total weight is equal to that of task two samples.

\paragraph{Extension of the three dimensional construction in \cref{thm: avg case counter example}  to higher dimensions.}
Fix an arbitrary orthonormal basis $\braces{\bfv_1, \dots, \bfv_d}$ and $\epsilon =0.4$. 
Set $\bfu = \epsilon \bfv_2 + \sqrt{1 - \epsilon^2} \bfv_3$ like the three dimensional construction. 
The first task spans the $d-1$ dimensional subspace $\lspan \paren{\braces{\bfv_1, \bfu, \bfv_4, \dots, \bfv_d}}$. 
As in the three dimensional construction, the second task is spanned by $\bfv_3$ only. 

In both the three dimensional and higher dimensional case $d-1$ samples from the first task would information theoretically be sufficient to learn $w^*$, but training until close to zero error might be challenging especially in the linear case. We experimentally verify this by directly solving the linear system and observing that replaying a few samples increases forgetting, while replaying $50$ samples will result in zero forgetting. 

\begin{figure}[H]
    \centering
    \includegraphics[width=0.6\linewidth]{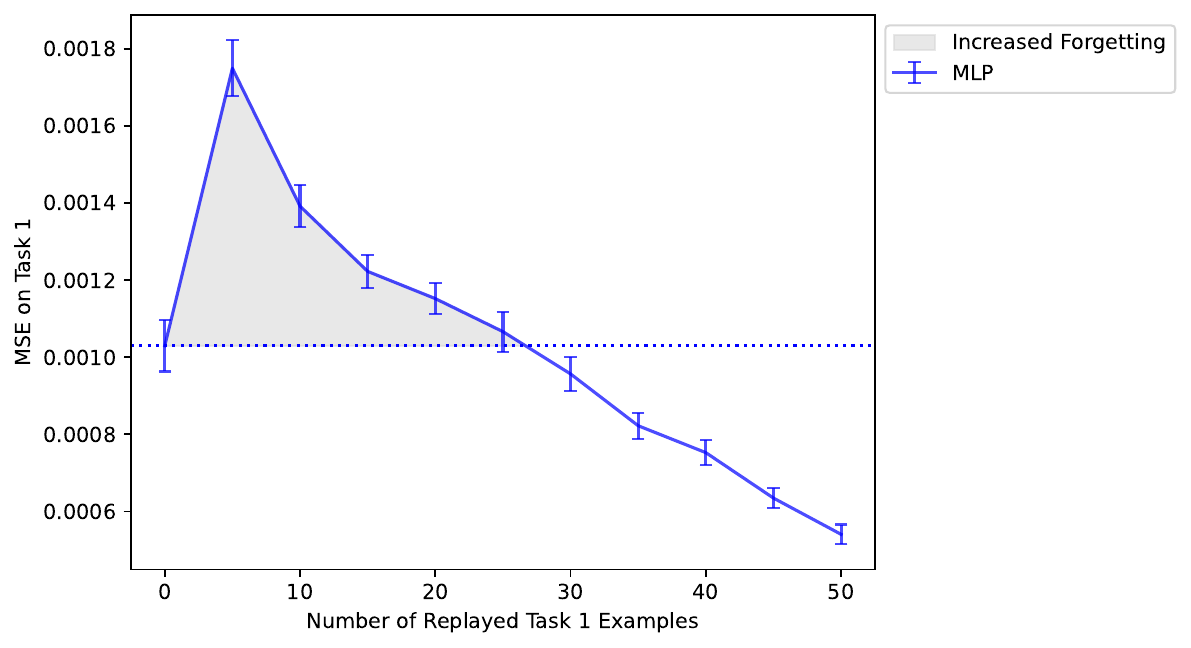}
    \caption{Same experiment as in \cref{fig: replay errors 50 dim } with a network of width $4 d $ instead}
    \label{fig: MLP smaller }
\end{figure}

\paragraph{Narrower Network}
We also include \cref{fig: MLP smaller }, 
which shows forgetting against the number of replay samples for a smaller network, where the width of the hidden layer is $4 d = 200$. The input data is generated with the same distributional parameters as in \cref{fig: replay errors 50 dim }. The training parameters for the second task were slightly different here.
Specifically, the exponential learning decay rate used on the second task was $0.9$. 

We note that it is possible that regularization, and training with small learning rate affect the observed pattern, especially with narrower networks.
However, studying the effect of regularization and hyper parameters on forgetting with replay is outside the scope of this paper.

\subsubsection{The effect of the angle between tasks while training with a MLP}\label{app: angles and mlp}
\begin{figure}[H]
    \centering
  \includegraphics[width=0.5\linewidth]{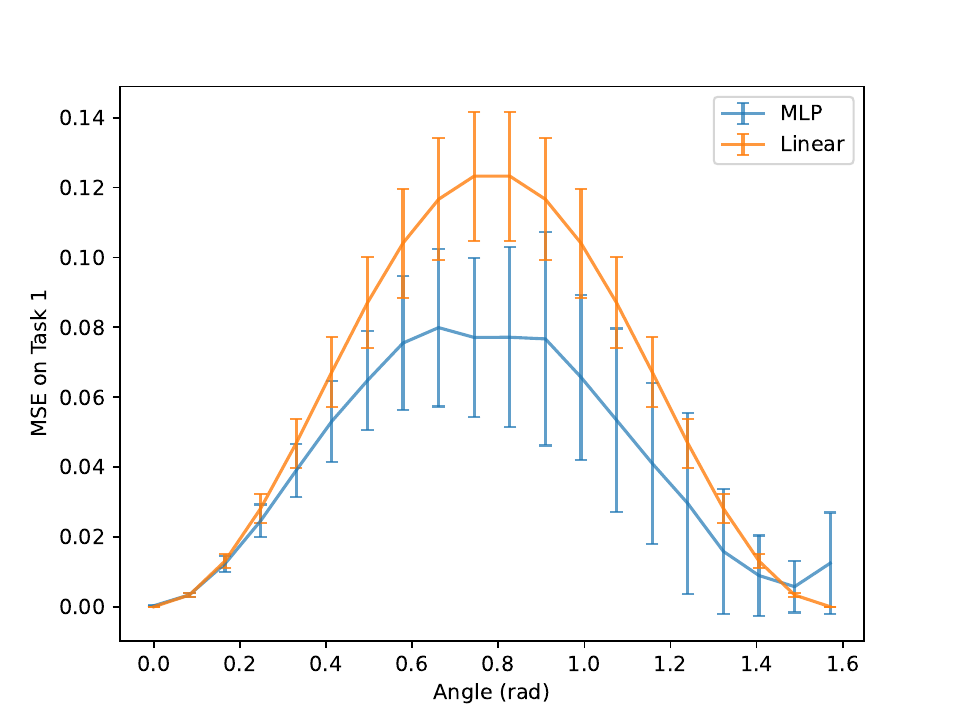}
    \caption{When training  using a MLP, the angle between task null spaces mostly has a similar effect on forgetting as the linear case, with  the exception of near orthogonal angles. Each point is averaged over 50 runs and the error bars here show standard deviation.}
    \label{fig: range of angles}
\end{figure}

We discussed in \cref{sec: average case results} how replay changes the angle between the two tasks in a way that increases forgetting on average. To understand whether a similar mechanism is responsible for the increase in forgetting due to replay in the nonlinear case,
we also look at the effect of the angle between two (linear) tasks on forgetting while using a nonlinear model for training. To do this, we pick the two tasks to be spanned by two $9$ dimensional subspaces in $\reals^{10}$, so that their null spaces are essentially given by two vectors $\bfu_1, \bfu_2$. 
We vary the angle between $\bfu_1$ and $\bfu_2$ and for each angle measure forgetting on the first task after training on the second task, see \cref{fig: range of angles}. 
We can see that forgetting of the MLP and linear model behave differently around angels that are close to $\pi/2$.
 In our three dimensional average case construction this won't matter, since initially without replay, the angle between the subspaces is close to zero, while with replay, the angle increases slightly but not a lot with high probability. 
Specifically, the construction is such that probability of replay leading to the angle being close to $\pi/2$ is very small.

Here we give more details on the experiments used to get \cref{fig: range of angles}.
The input distributions and what we referred to as the angle between two tasks has been already discussed in the last paragraph of \cref{sec:experiments}. The number of samples used per task is $100$. The linear model was trained using SGD with learning rate $0.1$. During training on the second task there was exponential decay rate of $0.95$.
The MLP had one hidden layer of width $128d$, and it was trained on the first task with starting learning rate $1\mathrm{e}-4$ and on the second task with initial learning rate $0.001$. in both cases (MLP training on task one and two), there was an exponential decay rate $0.6$. 
All the models for this experiment were trained for $5000$ epochs with the batch size $32$.

\subsection{Experiments on MNIST}
\label{appendix: experiments on MNIST}
In all the experiments, a fully connected network with two hidden layers of size $256$ was used. In all cases, training on each task was for $3$ epochs, with batch size of $32$, and using Adam \citep{Kingma2014AdamAM} with learning rate of $0.001$. 

\paragraph{Statistical tests.}
When we compared the means, we used Welch's t-test, which is similar to a student t-test while allowing the populations to have different variances. 

\paragraph{Rotated MNIST.} Rotated MNIST experiments are in a task incremental setting and use the training data for all the $10$ digits. The training data on the second task is the same as the training data on the first task, except that it is rotated. 
Forgetting is measure on test samples. 
The replay is done the same as the regression experiments. That is, for each class, a random sample is combined with the samples in the each batch and the replay samples are up-weighted such that the replay sample has equal weight to the rest of the samples.
The no replay baseline is what the literature might call the fine tuning baseline. The network is sequentially trained on the two tasks. The optimizer is reset after training on the first task. 

\paragraph{Split MNIST.} These experiments are in a class incremental setting, so the network had $4$ output heads. During evaluation on the first task, only the logits corresponding to the classes in the first task were taken into account. This is the case with or without replay. 
Again, the replay implementation here is similar to the regression experiments.

\subsection{Compute Resources}
\subsubsection{Regression Experiments}
The experiment in \cref{fig: range of angles}  took about 20 hours on a machine with single NVIDIA GeForce RTX 4080 GPU.
Each run of the experiments in figures \ref{fig: forgetting versus number of replayed samples} and \ref{fig: MLP smaller } would take about $0.5-1$ hour on a single NVIDIA A100-SXM4-80GB GPU. 
All the experiments did not use a significant amount of memory, since the input data was at most $50$ dimensional.

\subsection{MNIST Experiments} 
The experiments in \cref{fig: MNIST rotations} took about $6$ hours on a machine with a single NVIDIA GeForce RTX 4080 GPU for each rotation.
The experiment in \cref{fig: split MNIST} took about $2$ hours on the same machine.

\end{document}